%% file: main.tex
\documentclass{article}
\usepackage[a4paper, total={6.5in,8.5in}]{geometry}

\usepackage{subcaption}
\usepackage{graphicx}
\usepackage{algorithm}
\usepackage{algpseudocode}
\usepackage{booktabs} 
\usepackage{multirow} 
\usepackage[numbers, sort, comma, square]{natbib}
\usepackage{authblk}
\usepackage{wrapfig}
\usepackage{subcaption}
\usepackage{multicol}
\usepackage{array}
\usepackage{float}
\usepackage{hyperref}
\usepackage{booktabs}
\usepackage{siunitx}
\usepackage{subcaption}
\usepackage{rotating}
\usepackage{caption}
\usepackage{amsmath,amsfonts,bm}
\usepackage{amsmath}
\usepackage[dvipsnames]{xcolor}
\usepackage{enumitem}
\usepackage{placeins}
\usepackage{lscape}

\author[1]{Konstantin Donhauser\footnote{Equal contribution.}}
\author[1]{Javier Abad$^*$}
\author[2]{Neha Hulkund}
\author[1]{Fanny Yang}
\affil[1]{Department of Computer Science, ETH Zürich}
\affil[2]{MIT CSAIL}

\input{macros}

\title{Privacy-preserving data release leveraging optimal transport \\ and particle gradient descent}

\date{}

\begin{document}

\maketitle

\newcommand{\includeSubFigures}[1]{
            \begin{subfigure}{0.16\textwidth}
\includegraphics[width=\textwidth]{#1exp1_wdist_2_2Way_avg.pdf}
        \caption{$\swd_2^2$ dist.}
    \end{subfigure}
            \begin{subfigure}{0.16\textwidth}
\includegraphics[width=\textwidth]{#1exp1_l1_2Way_avg.pdf}
        \caption{TV dist.}
    \end{subfigure}
        \begin{subfigure}{0.16\textwidth}
\includegraphics[width=\textwidth]{#1exp1_synth_gradboost_test.pdf}
        \caption{downstream}
    \end{subfigure}
    \begin{subfigure}{0.16\textwidth}
        \centering
\includegraphics[width=\textwidth]{#1exp1_cov_fixed_frobenius_norm.pdf}
             \caption{cov. mat.}
        \end{subfigure}
        \begin{subfigure}{0.16\textwidth}
\includegraphics[width=\textwidth]{#1exp1_rand_coun_new_mean_dist.pdf}
             \caption{cout. queries}
    \end{subfigure}
        \begin{subfigure}{0.16\textwidth}
    \centering
\includegraphics[width=\textwidth]{#1exp1_rand_thrs_query_mean_dist.pdf}
             \caption{thresh. queries}
    \end{subfigure}
     \hspace{15mm} 
}

\newcommand{\includeTimeFigures}[2]{
            \begin{subfigure}{0.16\textwidth}
\includegraphics[width=\textwidth]{#1exp1_elapsed_time.pdf}
\caption*{#2}
\end{subfigure}
}


\newcommand{\includeBenchFigures}[1]{
    \centering
        \begin{subfigure}{\textwidth}
\includegraphics[width=\textwidth]{#11.pdf}
    \end{subfigure}\\
    \begin{subfigure}{\textwidth}
        \centering
\includegraphics[width=\textwidth]{#12.pdf}
        \end{subfigure}
       
}

\newcommand{\includeDistFigures}[1]{
    \centering
        \begin{subfigure}{0.47\textwidth}
\includegraphics[width=\textwidth]{#1wdist_1_2Way_avg.pdf}
    \end{subfigure}
    \begin{subfigure}{0.47\textwidth}
        \centering
\includegraphics[width=\textwidth]{#1newl1_2Way_avg.pdf}
        \end{subfigure}
       
}

\input{sections/intro}
\input{sections/background}

\input{sections/algo}

\input{sections/experiments}

\bibliographystyle{plainnat}
\bibliography{main}
\newpage
\appendix
\input{sections/appendix}

\end{document}

%% file: macros.tex
\usepackage[capitalize,noabbrev]{cleveref} 
\usepackage{mathtools}
\usepackage{amsfonts}
\newcommand{\dataset}{D}
\newcommand{\domain}{\mathcal X}
\newcommand{\algo}[1]{\mathcal A\left(#1\right)}

\newcommand{\swd}{\text{SW}}
\newcommand{\mcsamples}{N_{\text{MC}}}
\newcommand{\particles}{Z}

\newcommand{\queryloss}{\mathcal R}

\newcommand{\measuredmarginal}[1]{\hat \nu_{#1}}
\newcommand{\marginal}[1]{\nu_{#1}}
\newcommand{\marginalemb}[1]{\mu_{#1}}
\newcommand{\measuredempmarg}[1]{\hat \mu_{#1}}

\newtheorem{theorem}{Theorem}
\newtheorem{definition}[theorem]{Definition}

\newtheorem{lemma}{Lemma}
\newtheorem{hypothesis*}{Hypothesis}

\newif\ifshownotes
\shownotestrue

\newcommand{\embed}{\text{Emb}}
\newcommand{\domainP}{\Omega}

\newcommand{\dpdata}{D_{\text{DP}} }

\def\eqref#1{equation~\ref{#1}}
\def\Eqref#1{Equation~\ref{#1}}

\def\1{\bm{1}}

\DeclareMathAlphabet{\mathsfit}{\encodingdefault}{\sfdefault}{m}{sl}
\SetMathAlphabet{\mathsfit}{bold}{\encodingdefault}{\sfdefault}{bx}{n}

\def\gA{{\mathcal{A}}}

	[2021/11/17 Guillaume Wang's macros to write quick and dirty latex]

\newif\ifshownotes
\shownotestrue

\RequirePackage{xcolor}

\RequirePackage{xifthen}
\newcommand{\TODO}[1]{\ifshownotes%
    \ifthenelse{\isempty{#1}}%
    {\red{TODO}}%
    {\red{TODO: #1}}%
\else\fi}

\newcommand{\spnorm}[1]{{\left\vert\kern-0.25ex\left\vert\kern-0.25ex\left\vert #1 \right\vert\kern-0.25ex\right\vert\kern-0.25ex\right\vert}} 

\newcommand{\prob}[1]{\mathbb{P}\left(#1\right)}

\usepackage{bbm} 

\RequirePackage{xfrac}

%% file: sections/intro.tex
\begin{abstract}

We present a novel approach for differentially private data synthesis of protected tabular datasets, a relevant task in highly sensitive domains such as healthcare and government. Current state-of-the-art methods predominantly use marginal-based approaches, where a dataset is generated from private estimates of the marginals. In this paper, we introduce PrivPGD, a new generation method for marginal-based private data synthesis, leveraging tools from optimal transport and particle gradient descent. Our algorithm outperforms existing methods on a large range of datasets
while being highly scalable and offering the flexibility to incorporate additional domain-specific constraints.
\end{abstract}

\section{Introduction}

Distributing sensitive datasets is crucial for data-driven decision-making and methodological research in many fields, like healthcare and government. For example, epidemiologists and biostatisticians have relied on tools like SAS to analyze personal health data for decades \citep{dimaggio2013sas, dankar2012application}. However, there are significant concerns among patients about the use and disclosure of their information, especially among vulnerable groups~\citep{el2011case}.

Differential privacy (DP) has gained prominence as a vital approach to mitigate privacy concerns.   Its adoption extends well beyond theoretical frameworks, finding practical utility across industries and government organizations \citep{johnson2018towards,abowd2018us,aktay2020google}. 
In this paper, we target the problem of DP tabular data synthesis, a promising approach for creating high-quality copies of protected tabular datasets that adhere to privacy constraints. Any further task performed on these ``private" copies is thus guaranteed to comply with these constraints.

Numerous differential privacy methods have emerged to synthesize tabular datasets with privacy guarantees while preserving relevant statistics from the original dataset \citep{hu2023sok, tao2021benchmarking}. 
Marginal-based approaches are among the preferred methods, dominant in NIST challenges~\citep{mst} and consistently top-ranked in benchmarks~\citep{tao2021benchmarking}. These approaches select a set of marginals and perturb them in a DP-compliant manner. Subsequently, a synthetic dataset is \textit{generated} from these noisy marginals through a generation method.

In this paper, we introduce PrivPGD\footnote{See our GitHub repository for the source code: \url{https://github.com/jaabmar/private-pgd}.}, a novel DP-data generation method based on particle gradient descent \citep{chizat2018global, chizat2022sparse}. PrivPGD leverages an optimal transport-based divergence between the privatized and particle marginal distributions (Section~\ref{sec:slicedwd}) to integrate marginal information during gradient descent. This divergence can be efficiently approximated through parallel GPU processing, which is crucial for handling large datasets. Our approach has several important characteristics:

\begin{itemize}

\item \textit{State-of-the-Art performance.} PrivPGD outperforms state-of-the-art methods in a large benchmark comparison (9 datasets) across various metrics, including downstream task performance (Section~\ref{sec:experiments}).

  \item \textit{Scalability}. PrivPGD leverages a highly optimized gradient computation
 that can be parallelized on GPU, enabling the algorithm to efficiently construct large datasets with over 100,000 data points while accommodating many marginals, e.g., all 2-Way marginals.




\item \textit{Geometry preservation}. Many datasets contain features with inherent geometry, such as continuous features and some categorical features like age, which have rankings that should be retained in the synthetic data. Unlike state-of-the-art methods (see Section~\ref{sec:algorithms}), PrivPGD preserves this geometric structure, aligning more naturally with the nuances of real-world datasets.

\item \textit{Incorporation of domain-specific constraints.} As a gradient-based method, PrivPGD allows the inclusion of any additive penalization term to the loss function. This enables explicit enforcement of domain-specific constraints within the generation algorithm, providing a straightforward and efficient way to incorporate these requirements into the synthetic data. We provide examples and successful implementations of domain-specific constraints in Section~\ref{sec:domainspecific_exp}.

\end{itemize}

\subsection{Related work}
\label{sec:relatedwork}


In this section, we discuss related works on DP-data synthesis for tabular data and direct the readers to Section~\ref{sec:experiments} for extensive benchmark comparison. For a comprehensive overview, refer to the recent survey by \citet{hu2023sok}. We categorize the related work into marginal and query-based DP-data synthesis algorithms. Query-based algorithms handle a broad set of queries, while marginal-based algorithms focus on marginal queries and can be seen as a special case of query-based algorithms optimized for these queries.

\paragraph{Marginal-based algorithms} Marginal-based algorithms are prominent in the literature for private tabular data synthesis, achieving state-of-the-art performance on numerous benchmark tasks \citep{mst, tao2021benchmarking, hu2023sok}. These algorithms typically follow two main steps: selecting marginals \citep{cai2021data, mst, aim} and generating the dataset after adding noise to these marginals \citep{zhang2017privbayes, mckenna2019graphical, li2021dpsyn}. A common approach in data generation is PGM \citep{mckenna2019graphical}, used by winning methods in the NIST competitions~\citep{cai2021data, mst, aim}. However, PGM faces two major limitations: it is highly sensitive to the number of selected marginals, easily resulting in memory and runtime issues, and has limited capability in preserving domain-specific constraints. Other marginal-based generation methods include PrivBayes \citep{zhang2017privbayes}, which trains a Bayesian network, and Gradual Update Methods \citep{li2021dpsyn, zhang2021privsyn}, which iteratively update a random dataset to match the marginals.

\paragraph{Query-based algorithms} In contrast to marginal-based algorithms, query-based algorithms can handle a broader range of queries. Examples include DualQuery \citep{gaboardi14}, FEM \citep{vietri2020new}, RAP \citep{aydore2021differentially}, GEM \citep{liu2021iterative}, RAP++ \citep{vietri2022private}, and Private GSD \citep{pmlr-v202-liu23ag}. These methods construct a private dataset to answer a large set of predefined queries, often relying on gradient descent techniques \citep{aydore2021differentially, liu2021iterative}. Our method, PrivPGD, also uses particle gradient descent but does not require predefining a set of queries and is optimized to preserve marginal queries.


\paragraph{Other Algorithms} Generative Adversarial Networks (GANs) have been proposed for synthesizing private data \citep{xie2018differentially, jordon2019pate, torkzadehmahani2019dp}. Like other generator-based methods \citep{liu2021iterative, vero2023programmable}, GANs require specifying and fine-tuning a parametric model, which is challenging due to the high sensitivity to hyperparameters. In contrast, PrivPGD uses particles to represent data, eliminating the need for model selection and tuning, and is remarkably robust to hyperparameter variations. Moreover, GANs often fail to maintain basic distributional statistics for tabular datasets \citep{tao2021benchmarking}. DP-Sinkhorn \citep{cao2021don}, an optimal transport-based generative method, is not easily extendable to tabular data synthesis. Finally, copula-based approaches \citep{li2014differentially, asghar2019differentially, gambs2021growing} utilize Gaussian and vine copulas to model privatized marginal distributions, but these methods are computationally intensive.

%% file: sections/background.tex
\section{Preliminaries for differentially private data synthesis}
In this section, we summarize key concepts and introduce notation related to differentially private data synthesis used in the paper. 

In general, we consider our data to lie in a domain $\domain= \domain_1 \times  \cdots \times \domain_d$ that is discrete and has dimension $d$.
This assumption is not restrictive since the vast majority of DP-data synthesis algorithms for tabular data rely on a discretized version of the data even if it originally lies in a continuous domain\footnote{We refer the reader to \citep{zhang2016privtree} for a discussion on how to optimally discretize in a DP-way.}. In particular, we can represent every dimension as an integer in the discrete set $\domain_{i} = \{1, \cdots, k_i\}$ with $k_i \in \mathbb N_+$.

Differential privacy \citep{dp} is an algorithmic property that guarantees that individual information in the data is protected in the output of an algorithm; even when assuming that an adversary has access to the information of all other individuals in the dataset. We now provide the formal definition.

\begin{definition}
\label{eq:defdp}
    An algorithm $\gA$ is $(\epsilon,\delta)$-DP with $\epsilon>0$ and $\delta>0$ if for any datasets $\dataset, D'$
    differing in a single entry and any measurable subset $S \subset \mathrm{im}(\mathcal A)$ of the image of $\mathcal A$, we have
\begin{equation*}
  \prob{\algo{D}\in S} \leq \exp(\epsilon) \prob{\algo{D'}\in S} + \delta
\end{equation*}
\end{definition}

The goal of \emph{differentially private data synthesis} is to design a (randomized) algorithm $\gA$  that, for any dataset $\dataset \in \domain^n$ of size $n$, generates an output $\algo{D} \in \domain^m$ that is a differentially private ``copy" of $D$, potentially of different size $m \neq n$. 



\subsection{Marginal-based algorithms for private data synthesis }
\label{sec:marginal_based}
Our approach falls in the general category of marginal-based methods. They
 follow Algorithm~\ref{alg:mainalgo},  consisting of three steps: marginal selection, privatization, and generation.

\paragraph{Marginal selection} 
For any subset $S \subset \{1, \cdots, d\}$ of the dimensions,  we denote with $D_S \in \domain_S^n$ the dataset containing only the dimensions in $S$. For each subset $S$ we can define a corresponding marginal.
\begin{definition}
\label{def:marginal}
    We denote by $\marginal{S}[D] \in \mathcal P(\domain_S)$ the $S$-marginal of a dataset $D$, defined as the empirical measure of $D_S$ over the domain $\domain_S$.
    \end{definition}

In the first step of the algorithm, a set $\mathcal{S}$ of such subsets $S$ is selected, and equivalently a set of marginals. 
The problem of selecting marginals in a DP-way is an interesting problem on its own and has led to a significant amount of proposed methods~\citep{cai2021data, mst, aim, zhang2021privsyn}. 
In an
extension of Algorithm~\ref{alg:mainalgo}, sketched in Algorithm~\ref{alg:mwem}, the marginals are not all selected in the beginning, but sequentially chosen from 
a pre-defined pool of subsets $\mathcal S_W$ of $[d]$, often referred to as the workload.
More specifically, in every iteration $t$, we select the marginals with the largest total variation distance $\text{TV}\left( \marginal{S}[\dpdata^{(t-1)}], \marginal{S}[D]\right) $, where $\dpdata^{(t-1)}$ is the DP-dataset from the previous iteration $t-1$. This is done
in a DP-way using the exponential mechanism \cite{mcsherry2007mechanism}. Such approaches fall under the general MWEM~\citep{mwem,liu2021iterative} framework, which is the backbone of many DP synthesis methods. To control the overall privacy budget, the framework from Algorithm~\ref{alg:mwem} uses advanced compositional theorems.

\begin{figure}[!t]
\begin{minipage}[t]{.46\textwidth}
\noindent
\begin{algorithm}[H]
\caption{Standard data synthesis framework}\label{alg:mainalgo}
\begin{algorithmic}[1]
\Require Dataset $D$, privacy parameters $\epsilon$ and $\delta$
\State \textbf{select} set $\mathcal S$ of subsets $S$ of $\{1, \cdots, d\}$
\State \textbf{privatize} marginals $\marginal{S}[D]$ to obtain $(\epsilon, \delta)$-DP ``copies" $\measuredmarginal{S}$
\State \textbf{generate} data from privatized marginals $\measuredmarginal{S}$
\Statex\Return the DP dataset $\dpdata$
\end{algorithmic}
\end{algorithm}
\end{minipage}
\hfill
\begin{minipage}[t]{.52\textwidth}
\begin{algorithm}[H]
\caption{Extension to sequential query selection}\label{alg:mwem}
\begin{algorithmic}[1]
\Require Dataset \(D\), privacy parameters \(\epsilon\) and \(\delta\), workload \(\mathcal{S}_W\), rounds \(T\)
\For{\(t = 1, \dots, T\)}
    \State \textbf{select} \(S_t \in \mathcal{S}_W\)
    \State \textbf{privatize} the marginal \(\nu_{S_t}[D]\) to obtain the 
    \Statex\hspace{\algorithmicindent}\hspace{\algorithmicindent}DP-``copies" \(\hat{\nu}_{S_t}\)
    \State \textbf{generate} data from privatized marginals 
    \Statex\hspace{\algorithmicindent}\hspace{\algorithmicindent}\(\{\hat{\nu}_{S_j}\}_{j\leq t}\) to obtain \(D^{(t)}_{DP}\)
\EndFor
\State \Return DP dataset \(D^{(T)}_{DP}\).
\end{algorithmic}
\end{algorithm}
\end{minipage}
\end{figure}

\paragraph{Marginal privatization}
After a set of marginals has been selected, both frameworks contain a privatization and generation step.
A common choice for privatization is to apply the 
Gaussian mechanism \citep{mcsherry2007mechanism}, where we simply add i.i.d.~Gaussian noise to the empirical marginal $\marginal{S}[D]$. The variance $\sigma^2$ of the Gaussian depends on the privacy parameters $\epsilon$ and $\delta$. As a result, we obtain the signed measures  $\measuredmarginal{S}$:
\begin{equation*}
\label{eq:gaussian}
\forall x \in \domain_S:~~
\measuredmarginal{S}(\{x\}) = \marginal{S}[D](\{x\}) + \mathcal N(0, \sigma^2).
\end{equation*}

\paragraph{Data generation} 
Finally, in the last step, we generate a dataset from the noisy estimates of the marginals. 
For this purpose, existing methods 
typically aim to learn a distribution $\hat{p}$ that minimizes the squared loss:
\begin{equation*}
   \sum_{S \in \mathcal S, x \in \domain_S}  \left(\hat p_S(\{x\})  -  \measuredmarginal{S}(\{x\})\right)^2 
\end{equation*}
and then release the private synthetic  data $\dpdata$ by sampling from $\hat p$.
The predominant generation algorithm used by state-of-the-art methods \citep{mst, aim, cai2021data} is PGM~\citep{mckenna2019graphical}, which learns a graphical model using mirror descent. We refer to Section~\ref{sec:relatedwork} for further discussion.

\subsection{Sliced Wasserstein distance}
\label{sec:slicedwd}
The optimal transport literature offers a set of natural divergence measures that can be used for particle gradient descent, such as the Sinkhorn divergence or the Wasserstein distance. 
However, the computational complexity of these divergences scales at least quadratically (resp. cubically) with respect to the size of the support of the measures. This clearly defeats the purpose of a computationally efficient generation algorithm that can represent rich datasets with a large number of data points.
Instead, a widely used \citep{wu2019sliced, kolouri2018sliced, deshpande2018generative, deshpande2019max}, computationally efficient alternative is the squared sliced Wasserstein ($\swd_2^2$) distance \citep{bonneel2015sliced}; that is, the averaged squared Euclidean transportation over all $1$-dimensional projections. 

Formally, for $\theta\in\mathbb{R}$, let $g^\theta(x) = \langle x, \theta\rangle$ and $g^\theta_{\#}$ be the pullback measure induced by $g^\theta$. Moreover, let $\lambda^p$ be the uniform distribution over the sphere $\mathcal S^{p-1}$. For any Euclidean subspace $\Omega \subset \mathbb{R}^p$ and probability measures $\mu, \nu \in \mathcal{P}(\Omega)$ with support $\Omega$,  the $q$-SW distance is defined as follows:
\begin{equation}
\label{eq:def_sliced_wasserstein}
    \swd_q(\mu, \nu) =  \left(\int_{\mathcal S^{p -1}} W_q^q(g^\theta_{\#} \mu, g^\theta_{\#}\nu) d\lambda(\theta) \right)^{1/q}, 
\end{equation}
where 
$W_q^q$ is the $q$-th power of the $q$-Wasserstein distance for $q \geq 1$. The Wasserstein distance 
over $1$-dimensional distributions, as it appears in \Cref{eq:def_sliced_wasserstein},
has a well-known closed-form expression:
\begin{equation}
\label{eq:def_wasserstein}
    W_q(g^\theta_{\#} \mu, g^\theta_{\#}\nu)  = \left(\int_u \vert F^{-1}_{g^\theta_{\#} \mu}(u) - F^{-1}_{ g^\theta_{\#}\nu }(u)\vert^q du\right)^{1/q}, 
\end{equation}
 where $F_{g^\theta_{\#} \mu}(u)$ (resp.~$ F_{ g^\theta_{\#}\nu }(u)$) represents the cumulative  function and $F^{-1}$ its inverse. Finally, in the special case of $q=1$, we have the  following identity for the $1$-Wasserstein distance by Vallender \citep{vallander}, which we will later leverage in our algorithm:\begin{equation}\label{eq:vallander}
    W_1(g^\theta_{\#} \mu, g^\theta_{\#}\nu) = \| F_{g^\theta_{\#} \mu}(u) - F_{ g^\theta_{\#}\nu }(u)\|_{L_1(\mathbb R)}.
\end{equation}
Importantly, this identity allows us to naturally extend the definition of the (sliced) $1$-Wasserstein distance to signed measures, as demonstrated by \citet{boedihardjo2022private}.

%% file: sections/algo.tex
\section{PrivPGD: a particle gradient descent-based generation method}

We introduce \textit{PrivPGD} (Algorithm~\ref{alg:pgd}), a novel approach for solving the generation step in marginal-based tabular data synthesis (Algorithm~\ref{alg:mainalgo}).
Unlike other marginal-based methods \citep{zhang2017privbayes,mckenna2019graphical}, PrivPGD does not construct a dataset by sampling from a learned distribution. Instead, it directly propagates particles in an embedding space to minimize the sliced Wasserstein distance. A distinct advantage is that, through particle gradient descent, we can easily enforce domain-specific constraints by adding a penalization term $\hat \queryloss$ to the loss. A similar approach for incorporating domain-specific constraints has recently been proposed for generator-based data synthesis algorithms \citep{vero2023programmable}, where the authors provide various examples of such losses. 

In summary, our data generation method returns a DP-dataset $\dpdata$ given the following inputs:
\begin{enumerate}
\item A set of differentially private finite signed measures {$ \{\measuredmarginal{S}\}_{ S \in \mathcal S}$}  constructed as in Algorithm~\ref{alg:mainalgo}.
    \item  A differentially private regularization loss $\hat \queryloss$ incorporating domain-specific constraints.
    \end{enumerate}


\subsection{Preliminaries: Embedding}
\label{sec:embedding}
PrivPGD crucially relies on an embedding ${\embed: \domain \to \domainP}$ of the (discretized) domain $\domain$ into a compact Euclidean product space $\domainP= \domainP_1 \times \cdots \times \domainP_d$.
We simply choose $\domainP = [0,1]^d$ and map every $x \in \domain$ to equally-spaced centers
\begin{equation}
\label{eq:embedding}
    \embed(x)_i = \frac{ 2 x_i -  1}{2 k_i} \in [0,1].
\end{equation}
This choice of embedding preserves the order in $\domain$, which is essential for variables such as age or any discretized continuous variables. In line with common practices in the literature \citep{mst,tao2021benchmarking}, we discretize continuous data using equally spaced bins. This method ensures that the embedding accurately represents the scaled distances between the centers.

We acknowledge that features like race, where imposing an ordering might be inappropriate, could be embedded into the space $[0,1]^2$ so that the centers are equidistant. Similarly, when embedding categorical variables representing locations, an embedding that preserves geographical distances might be preferable. While it would be interesting to explore other embeddings, we leave it to future work.


\paragraph{Particles} PrivPGD aims to construct $m$ data points in the embedding space, ensuring their empirical distribution closely approximates the projection of the privatized signed measure $\hat \nu_S $. For any set of points $\particles \in \domainP^m$, which we also refer to as the $m$ \textit{particles}, we define $\particles_S \in \Omega_S^m$ as the projection of these particles onto the embedding $\Omega_S$ of $\domain_S$.
  Moreover, we define $S$-marginals over $\Omega$ as:
\begin{definition}
    We denote by $\marginalemb{S}[\particles] \in \mathcal P(\domainP_S)$ the $S$-marginal of  the particles $\particles$, defined as the empirical measure of $\particles_S$ over the domain $\domainP_S$.
\end{definition}

\begin{figure}[!t]
  \begin{algorithm}[H]
\caption{\textbf{Priv}ate \textbf{P}article \textbf{G}radient \textbf{D}escent }\label{alg:pgd}
\begin{algorithmic}[1]
\Require  DP marginals $ \{\measuredmarginal{S}\}_{ S \in \mathcal S}$, regularizer  $\hat \queryloss$ , number of particles $m$ 

\State \textbf{projection:} $\forall S \in \mathcal S$, construct  the empirical measures $ \measuredempmarg{S}$ from $ \measuredmarginal{S}$ 

 \State \textbf{optimization:} randomly initialize $\particles^{(0)} \in \Omega^m$

\For{$t = 1, \cdots, T$}
\State \textit{select} batch $\mathcal S_{\text{batch}} \subset \mathcal S$ 
\State \textit{compute} the gradient at $\particles^{(t-1)}$ of 
       $\sum_{S \in \mathcal S_{\text{batch}}} \swd_2^2(\marginalemb{S}[\particles],  \measuredempmarg{S}) + \lambda \hat \queryloss(\particles)$

\State \textit{update} $Z^{(t)}$ using any first order optimizer

\EndFor
 \State \textbf{finalization step:} construct $\dpdata$ from $\particles^{(T)}$
 \Statex \Return $\dpdata$
\end{algorithmic}
\end{algorithm}
\vspace{-0.3in}
\end{figure}

\subsection{Projection step}
\label{sec:projection}

The preliminary embedding step allows us to define the particles $Z$ within a convenient domain $\domainP$, which we choose to be the hypercube with a fixed grid. In the projection step, the goal is to transform the privatized signed measure $\measuredmarginal{S}$ into a proper probability measure $\measuredempmarg{S}$ that can be ``plugged into" the Wasserstein distance. Further, since we aim to find particles whose empirical distribution closely approximates the signed measure, we quantize $\measuredempmarg{S}$ using the same number of particles, $m$.

\paragraph{Projection} First, note that the embedding $\embed$ from  \Cref{eq:embedding} defines corresponding finite signed measures $\hat \omega_S$ over $\embed(\domain)_S \subset \domainP_S$ for each privatized signed measure $\measuredmarginal{S}$. By default, the sliced Wasserstein distance that we minimize in the optimization step (\Cref{subsec:optimization}) is defined for probability measures. For $q=1$ we can extend the sliced 1-Wasserstein distance from~\Cref{eq:vallander} to signed measures and obtain a probability measure.
Inspired by \citet{boedihardjo2022private} (see also \citep{donhauser2023certified}), we transform 
by solving 
\begin{equation}
    \hat \omega_{S, \mathbb P} = \arg\min_{w \in \mathcal P(\embed(\domain)_S) } \swd_1( w, \hat \omega_S).
    \label{eq:objective_preprocessing}
\end{equation}
We solve this convex optimization problem  using gradient descent. We also approximate the integral in $\swd_1$ using Monte Carlo samples. Importantly, minimizing the objective in~\Cref{eq:objective_preprocessing} allows for preserving the geometry from the signed measures in the probability measures.

\paragraph{Quantization} We further quantize the finite probability measures $\hat \omega_{S, \mathbb P}$ using $m$ particles 
$\hat \particles_S \in \Omega^m_S$, i.e., 
\begin{equation}
    \measuredempmarg{S} = \frac{1}{m} \sum_{i=1}^m \delta[\hat \particles_S^i],
    \label{eq:quant}
\end{equation}
such that $\hat \mu_S \approx \hat \omega_{S, \mathbb P}$. 
  This can be achieved by using any standard quantization technique with a negligible error for a sufficiently large number of particles.
  We apply the quantization in \Cref{eq:quant} with $m$ particles, thus ensuring that both $\marginalemb{S}[\particles]$ and $  \measuredempmarg{S} $ are empirical measures over the same number of particles. 
  
\subsection{Optimization and finalization step}

\label{subsec:optimization}
In the optimization step, we aim to generate a dataset by finding particles $Z \in \domainP^m$  that are close to the differentially private measure $\measuredempmarg{S}$ constructed in the projection step.
In particular, the final particles should minimize the squared sliced Wasserstein distance $\swd_2^2$ (\Cref{eq:def_sliced_wasserstein}) between the empirical marginal distributions of the particles $\marginalemb{S}[\particles]$ and $\measuredempmarg{S}$
We additionally incorporate domain-specific constraints via a DP differentiable penalty term $\hat \queryloss: \Omega^m \to \mathbb R$. 
Formally, for a regularization  strength $\lambda$, we
 run mini-batch particle gradient descent on
\begin{equation}
\label{eq:loss_gd}
   \mathcal L_{\mathcal S}(\particles) :=  \sum_{S \in \mathcal S} \swd_2^2(\marginalemb{S}[\particles], \measuredempmarg{S} )+ \lambda \hat \queryloss(\particles),
\end{equation}
where $\swd_2^2$ is the squared sliced Wasserstein distance.  
\paragraph{Computing the gradient of the $\swd_2^2$ distance} 
For computing the gradient, we leverage the
fact that the 1-dimensional $2$-Wasserstein distance between $g^\theta_{\#}\marginalemb{S}[\particles]= \frac{1}{m} \sum_i \delta(y_i)$ and $g^\theta_{\#} \measuredempmarg{S} = \frac{1}{m} \sum_i \delta(y'_i)$ with $y_i, y'_i \in \mathbb R$ has a closed-form expression
\begin{equation}
\label{eq:wasserstein_simple}
     W_2^2(g^\theta_{\#}\marginalemb{S}[\particles],g^\theta_{\#} \measuredempmarg{S})  = \frac{1}{m} \sum_i (y_{[i]}- y'_{[i]})^2,  
\end{equation}
where $y_{[i]}$ (resp.~$y'_{[i]}$) denotes the $i$-th largest element. 
\Cref{eq:wasserstein_simple} and its gradient can be computed efficiently by running a sorting algorithm, which is parallelizable on modern GPU architectures. We then approximate the $\swd_2^2$ using $\mcsamples$ Monte Carlo samples for $\theta$. Consequently, we achieve a runtime complexity of $O(\vert \mathcal S \vert \cdot \mcsamples \cdot m \log m)$ for obtaining the gradient of the first term in \Cref{eq:loss_gd}.

\paragraph{Finalization step} Finally, after running particle gradient descent for $T$ iterations, we obtain the dataset $\dpdata \in \domain^m$ by mapping every final particle in $\particles^{(T)}$ to the closest point in $\embed(\domain) \subset \domainP$.

\subsection{Privacy guarantees}
We outline the privacy guarantees for the dataset $\dpdata$ generated using PrivPGD. Leveraging the data post-processing property \citep{gauss_mechanism}, we establish the following lemma:
\begin{lemma}
    Assume that the noisy marginals $ \{\measuredmarginal{S}\}_{ S \in \mathcal S}$  and the regularization loss $\hat \queryloss$ are generated from two independent $(\epsilon_1, \delta_1)$- and $(\epsilon_2, \delta_2)$-DP mechanisms. Then, the output of Algorithm~\ref{alg:pgd},   $\dpdata$, is $(\epsilon_1 + \epsilon_2, \delta_1+ \delta_2)$-DP.
\end{lemma}
Following standard practices in the literature \citep{mst, aim, vietri2022private}, we employ the Gaussian mechanism to generate both the noisy marginals and the (parameterized) regularization loss for the experiments described in the next section. This approach ensures that the synthesized dataset $\dpdata$ adheres to differential privacy guarantees.

%% file: sections/experiments.tex
\section{Experiments}
\label{sec:experiments}
In this section, we present a systematic large-scale experimental evaluation of our algorithm.
\subsection{Experimental setting}
\label{subsec:expsetting}
\paragraph{Dataset} 
We use $9$ real-world datasets from various sources, detailed in Appendix~\ref{appendix:datasets}. Each dataset contains no fewer than $50,000$ data points, ranges from $3$ to $22$ dimensions and is linked to a binary classification or regression task, which we use to evaluate the downstream error. For these evaluations, we allocate $80\%$ of the data as private data $D$ and use the remaining $20\%$ for test data $D_{test}$.  We discretize every dimension containing real values or integer values exceeding a range of $32$ into $32$ equally-sized bins.

\paragraph{Privacy Budget} We use $\epsilon =2.5$ and $\delta =10^{-5}$ as default choices. According to the National Institute of Standards and Technology (NIST), $\epsilon$ values below $5$ can be considered as strong privacy protection and real-world applications commonly use values above $2.5$ \citep{NIST-diff-privacy}.

\paragraph{Metrics} We evaluate the statistical and downstream task performance of our algorithm with the following standard metrics for DP data  synthesis:
\begin{itemize}

    \item[1.] \textit{downstream error:} The classification/regression test error, i.e., the $0$-$1$ (resp. mean squared) error, of gradient boosting trained on the synthesized data $\dpdata$  and evaluated on the (non-privatized) test dataset $D_{test}$. 
    \item[2.] \textit{covariance error:} The Frobenius norm of the differences of the centered covariance matrices of $\embed(D) $ and $\embed(\dpdata)$ divided by the Frobenius norm of the centered covariance matrix of $\embed(\dpdata)$. Since the embedding just rescales the (discretized) variables, this is equivalent to computing the covariance matrix error of the normalized data, up to a constant.
\end{itemize}

Moreover, we use the relative average over $J =200$  query differences between $\dpdata$ and $D$
    \begin{equation}
    \label{eq:countquery}
      \frac{ \frac{1}{J}\sum_j\vert \text{query}_j(\dpdata) - \text{query}_j(D)\vert }{  \frac{1}{J}\sum_j \text{query}_j(D)}.
    \end{equation}

We instantiate the query difference 
for two commonly used queries (see e.g., \cite{vietri2022private}):
    \begin{itemize}
        \item[3.] \textit{count. queries:} $3$-sparse counting queries 
with $\text{query}_j(D)= \text{count}_j(D) = \frac{1}{n} \sum_i 
\mathbbm{1}[x_i \in \mathcal A_j] $  with  $A_j$ the full hypercube $\forall l: A_j^l = {1, \dots, k_l}$ except for $3$ random dimensions where the $A_j^l$ is an interval with uniformly drawn lower bound and subsequently drawn upper bound. We use rejection sampling to ensure that at least 5\% and at most 95\% of the samples of the original dataset fall in this interval for every $j$. 
\item[4.] \textit{thresh. queries:}
    $3$-sparse linear thresholding queries 
with $\text{query}_j(D)=\text{thrs}_j(D) = \frac{1}{n} \sum_i  
\mathbbm{1}_{\langle x_i, \theta \rangle + b_j >0} $; $\theta$ is a random $3$-sparse direction and $b_j$ is uniformly drawn from the interval $[\min_{x \in D} \langle x, \theta \rangle,\max_{x \in D} \langle x, \theta \rangle]$.
 \end{itemize}

 Finally, we compute the average sliced Wasserstein distance and the total variation distance. The former is approximately minimized by PrivPGD while the latter by existing marginal-based approaches (e.g., \citep{aim,mst}):
\begin{itemize}
    \item[5.] \emph{average $\swd_1$ dist.:}  
    The average $\swd_1$ distance over all $2$-Way marginals between the embedded empirical probability measures  of the original dataset $D$  and the DP dataset $\dpdata$;
    $\mu_S[D], \mu_S[\dpdata] \in  \mathcal P(\embed(\domain)_S)$: 
    \begin{equation}
        {d \choose 2}^{-1} \sum_{S \subset [d]; \vert S \vert =2 } \swd_1 \left( \mu_S[D] , \mu_S[\dpdata]\right) 
    \end{equation}
        \item[6.] \emph{average TV dist.:} The average total variation distance $\text{TV}$ over all $2$-Way marginals between the original dataset $D$ and the DP dataset $\dpdata$
    \begin{equation}
        {d \choose 2}^{-1} \sum_{S \subset [d]; \vert S \vert =2 } \text{TV}\left( \marginal{S}[D], \marginal{S}[\dpdata]\right) 
    \end{equation}
\end{itemize}

  \paragraph{Implementation of PrivPGD}
 We implement PrivPGD using \textit{PyTorch} on a GPU and use the same hyperparameters for all experiments. We select all $2$-Way marginals and use the Gaussian mechanism to construct DP-copies of them  (step 2 in Algorithm~\ref{alg:mainalgo}). We then generate a dataset by running PrivPGD (Algorithm~\ref{alg:pgd}) with  $100k$ particles.

 For the \emph{projection step} (Section~\ref{sec:projection}), we use $200$ MC random projections to approximate the $\swd_1$ distance. We construct the finite measure $ \hat \omega_{S, \mathbb P}$ by running gradient descent for $1750$ iterations using \text{Adam} with an initial learning rate of $0.1$ and a linear learning rate scheduler with step size $100$ and multiplicative factor $0.8$. As initialization, we use the probability measure obtained when setting all negative weights of $\hat \omega_{S}$ to zero and subsequently normalize the positive finite measure.

In the \emph{optimization step} (Section~\ref{subsec:optimization}), we  approximate the $\swd^2_2$ using $N_{\text{MC}} =10$ projections. We minimize the objective in \Cref{eq:loss_gd} by running gradient descent for $1000$ epochs (where in every epoch every marginal is seen exactly once) using an initial learning rate of $0.1$ and a linear learning rate scheduler with step size $50$ and multiplicative factor $0.75$. We divide $\mathcal S$ into mini-batches of size $5$ and randomly set $80\%$ of the gradient entries to zero. We use \textit{Sparse Adam} from the $\textit{PyTorch}$ package.

\paragraph{Implementation of the metrics} We use the implementation from \textit{scikit-learn}~\citep{sklearn_api} for Gradient Boosting using the standard hyperparameters. Since discretization is a separate problem on its own, we use the discretized dataset in all experiments. 

\subsection{Algorithms for benchmarking}
\label{sec:algorithms}
We benchmark PrivPGD against a number of baselines including representative PGM-based and query-based methods. The PGM-based methods include MST \citep{mst} and AIM \cite{aim}. For AIM, we choose all $2$-Way marginals as workload. The following query-based algorithms include Private GSD \citep{pmlr-v202-liu23ag}, RAP \cite{aydore2021differentially} and GEM \citep{liu2021iterative}, where we choose all 2-Way marginals as queries. 

\emph{Implementation of AIM and MST.} We implement MST and AIM using the code provided by the authors\footnote{ \url{https://github.com/ryan112358/private-pgm}.}. We choose the initial learning rate to be $1.0$ and run mirror descent for $3000$ iterations. We fix the hyperparameters for all experiments. Moreover, we slightly modify the code for MST and AIM by increasing the sensitivity used in the Gaussian \citep{gauss_mechanism} and exponential mechanisms \citep{mcsherry2007mechanism} to give accurate privacy guarantees for the differential privacy model from Definition~\ref{eq:defdp}. 
We refer to \citep{aim} for an overview of both mechanisms in the context of DP data release. 

\emph{Implementation of query-based algorithms.} We use the one-shot version of Private GSD from the paper with an elite size of 2 and early stopping, and tune the number of generations $G \in \{200k, 500k\}$, the mutation and crossover populations $P_{mut} = P_{cross} \in \{50, 100, 150, 200, 500\}$, and the number of synthesized data points $m \in \{2k, 100k\}$. For GEM, we keep the default hyperparameters, tuning the number of iterations $T \in \{3, 10, 30, 50, 100, 150, 200\}$ and the model architecture $layers_{MLP} \in \{[512, 512, 1024], [128, 256]\}$. 
For RAP, we also keep the default hyperparameters and tune  $T \in \{3, 10, 30, 50, 100, 150, 200\}$, the learning rate $lr \in \{0.0001, 0.001, 0.03\}$ and $m \in \{2k, 500k\}$. This hyperparameter optimization includes the configurations considered by \citet{pmlr-v202-liu23ag}. All methods are implemented using their official versions, with hyperparameters fine-tuned for each dataset. GEM and RAP could not be run for the ACS Employment dataset due to exceeding our memory constraint of 20 GB.

 \begin{figure*}[t!]

    \centering
  \begin{subfigure}{\linewidth}
        \centering
\includegraphics[width=0.9\textwidth]{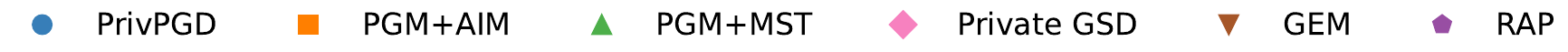}
    \end{subfigure}
    \includeBenchFigures{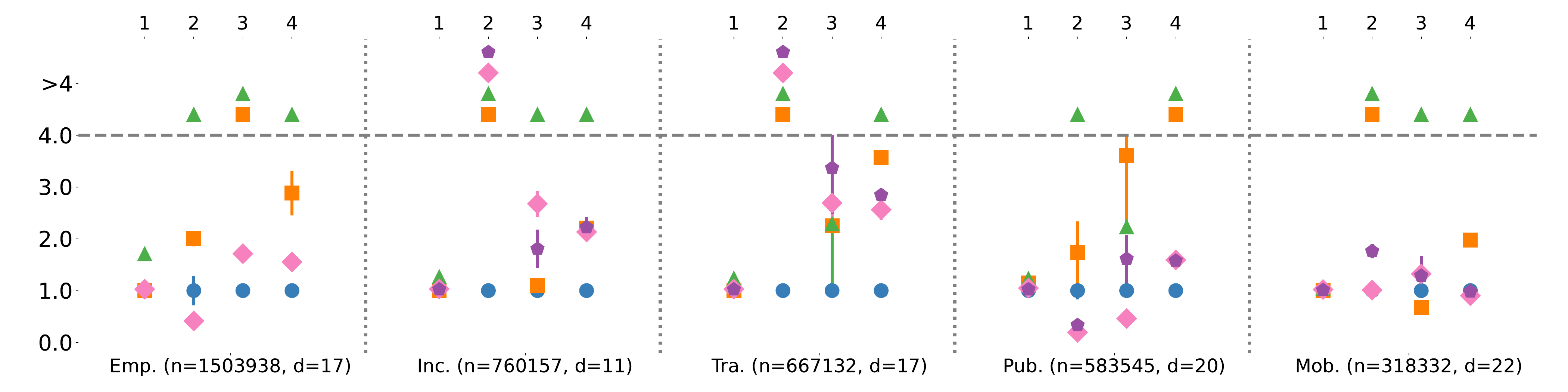}
        \centering
    \caption{
    \small 
    Comparison of PrivPGD with all $2$-Way marginals against state-of-the-art methods based on metrics from Section~\ref{subsec:expsetting}: 1) downstream error, 2) covariance error, 3) count. queries error, and 4) thresh. queries error, across 9 tabular datasets. For each method, we plot the $\log_2$ ratio of the errors, using PrivPGD's average error as the denominator, and report the mean and standard deviation over 5 runs. We cut at a log ratio of $y=4$ (dashed line) and list all methods exceeding this threshold above this line in order. We set $\epsilon = 2.5$ and $\delta = 10^{-5}$.
}    
    \label{fig:comp}
    \vspace{-0.2in}
\end{figure*}

  \subsection{Comparison with baselines}
\label{sec:exp_priv_marginals}

We now present how the performance of PrivPGD compares against the state-of-the-art algorithms for DP data synthesis. Figure~\ref{fig:comp} illustrates the relative performance of PrivPGD compared to other methods, using metrics from Section~\ref{subsec:expsetting}. PrivPGD consistently ranks as either the best or the second-best in most metrics and datasets, with a few exceptions such as the covariance error in the ACS Public Coverage dataset.

\paragraph{Comparison with PGM}
PrivPGD systematically outperforms both variants of PGM, which are the state-of-the-art for marginal-based tabular data synthesis, often by a significant margin. Specifically, it is better than PGM+MST in covariance and query errors across datasets, except for thresholding query errors in the Diabetes dataset, and performs at least as well as PGM+AIM, usually surpassing it, with the notable exception of thresholding query errors in the ACS Mobility dataset. For downstream tasks, PrivPGD performs comparably to PGM+AIM and consistently better than PGM+MST.

\paragraph{Comparison with query-based algorithms}
Furthermore, PrivPGD is also the preferred method in most scenarios when compared to query-based approaches, especially against GEM and RAP. While Private GSD provides competitive performance and occasionally surpasses PrivPGD – for instance, in the ACS Public Coverage dataset – PrivPGD usually emerges as the best-performing method. On many datasets it significantly outperforms Private GSD in all metrics, as exemplified by the Taxi, Black Friday, and ACS Traveltime datasets.

\begin{figure*}
    \centering
    \begin{subfigure}{\linewidth}
        \centering
        \includegraphics[width=0.9\textwidth]{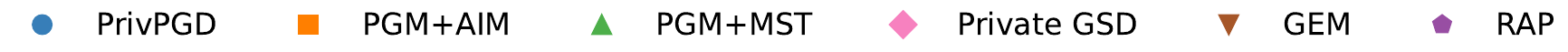}
    \end{subfigure}
    \includeDistFigures{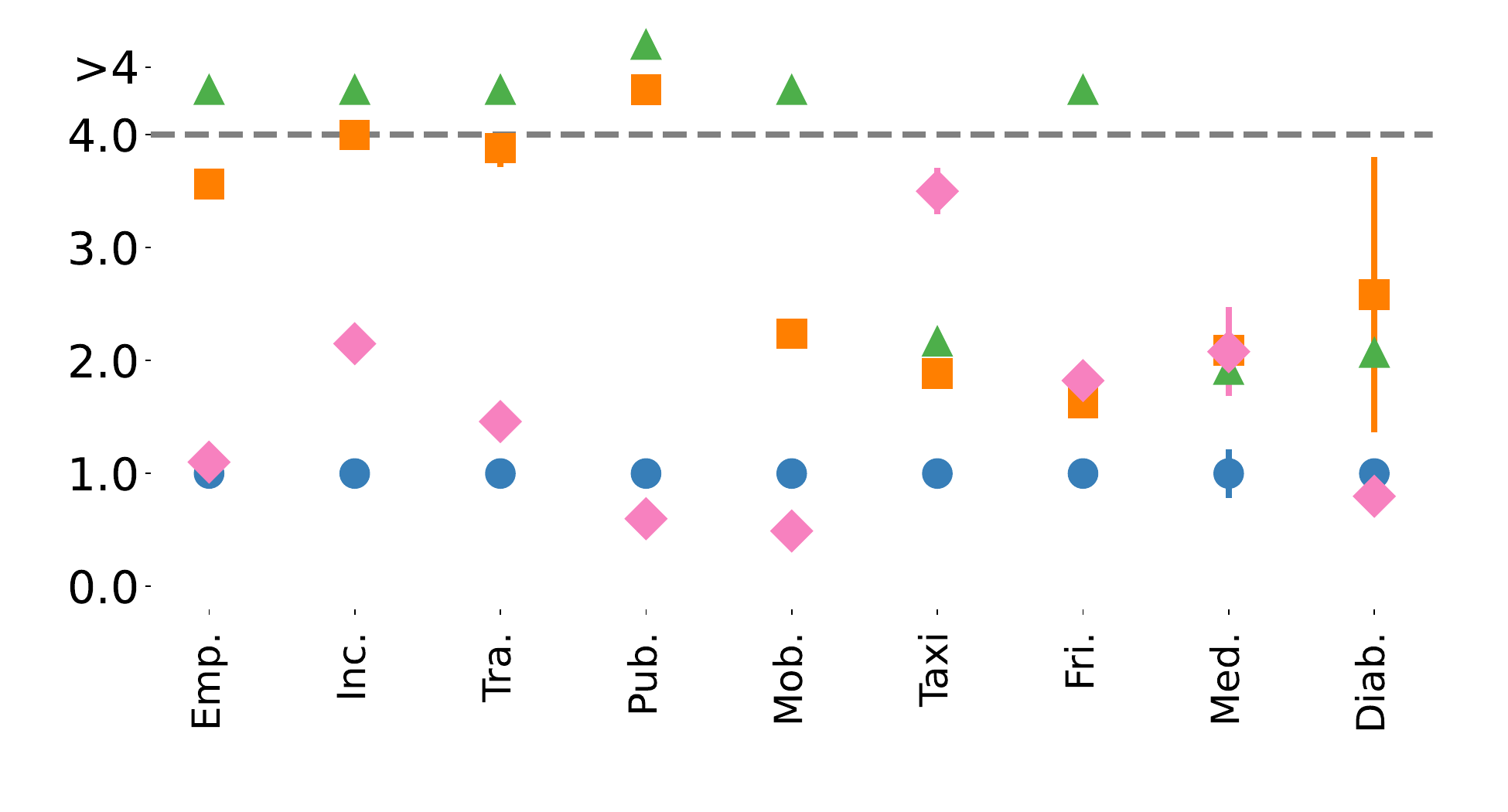}
        \centering
        \caption{\small Comparison of average $\swd_1$ distance (left) and average TV distance (right) for PrivPGD against state-of-the-art methods across 9 tabular datasets. Similar to Figure~\ref{fig:comp}, we report the mean and standard deviation (5 runs) of the $\log_2$ ratio of errors. We set $\epsilon = 2.5$ and $\delta = 10^{-5}$.}

        \label{fig:comp_sw_tv_2.5}
\end{figure*}

\paragraph{$\swd_1$ and TV distance} Finally, Figure~\ref{fig:comp_sw_tv_2.5} illustrates a noticeable performance gap between PrivPGD and other methods when comparing the average sliced Wasserstein distance and the total variation distance. PrivPGD effectively minimizes the former, while state-of-the-art methods like PGM primarily target the latter. Our experiments demonstrate the advantage of minimizing a geometry-aware loss function like the sliced Wasserstein distance over the total variation distance.

\subsection{Enforcing additional domain-specific constraints}
\label{sec:domainspecific_exp}

\begin{wrapfigure}{r}{0.52\textwidth}
\vspace{-0.3in}
    \centering
        \begin{subfigure}{\linewidth}
        \centering
\includegraphics[width=\textwidth]{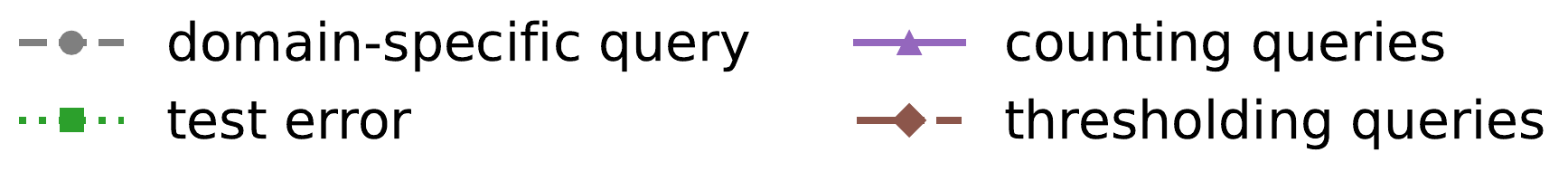}
    \end{subfigure}
    \begin{subfigure}{.45\linewidth}
        \centering
        \includegraphics[width=\textwidth]{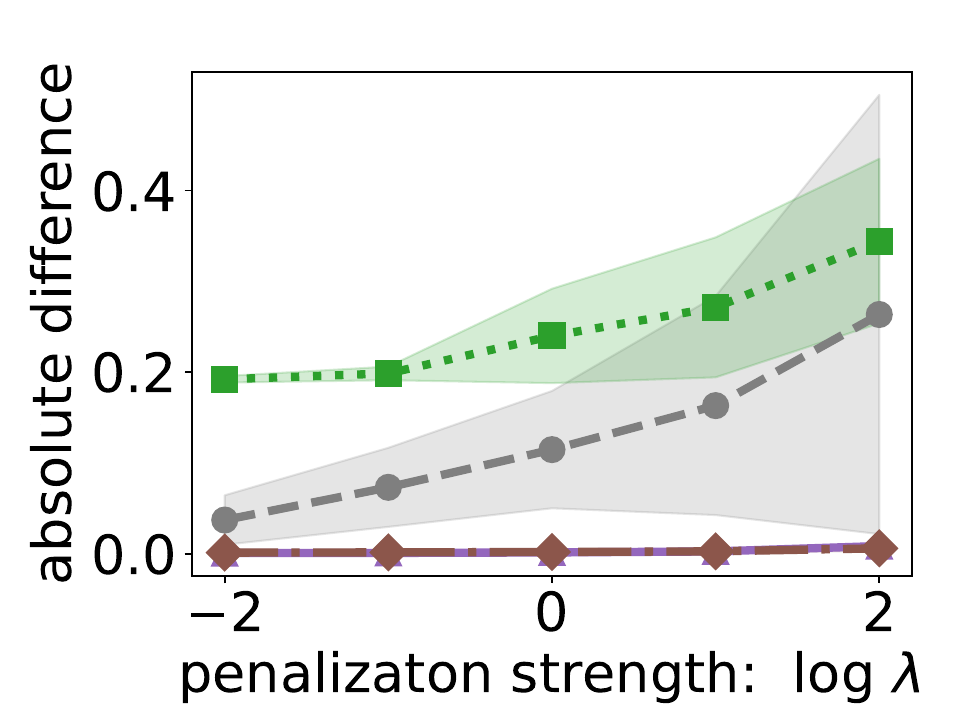}
        \caption{ACS Income}
    \end{subfigure}%
    \begin{subfigure}{.45\linewidth}
        \centering
        \includegraphics[width=\textwidth]{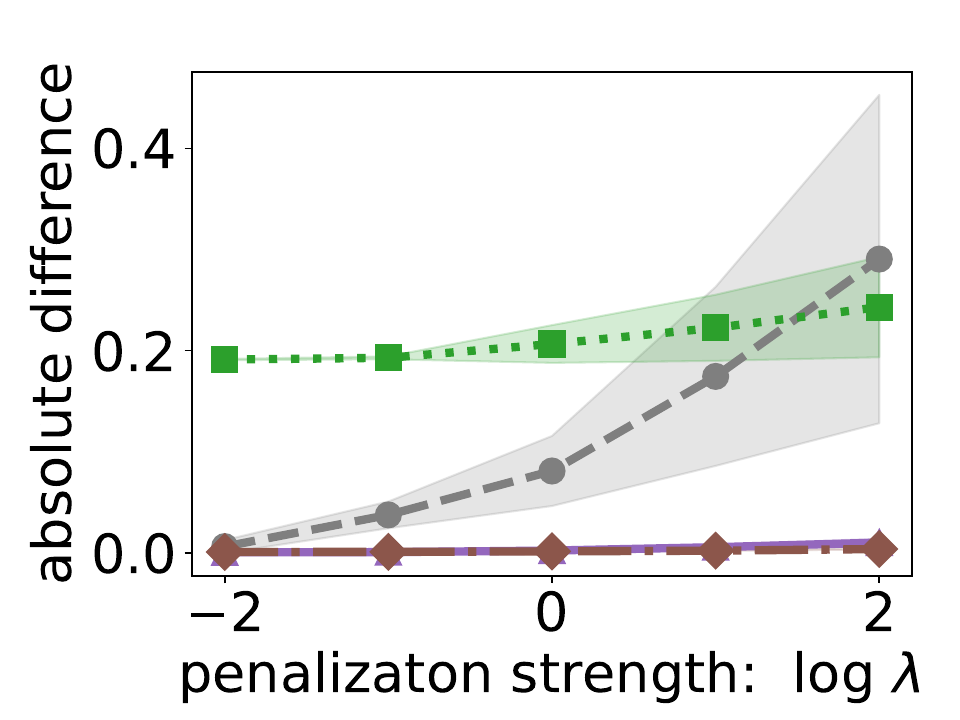}
        \caption{ACS Employment}
    \end{subfigure}
    \caption{\small  The absolute error of the domain-specific query (larger is better), the downstream classification error (smaller is better), and the absolute error over counting and thresholding queries (smaller is better), i.e., only the numerator in ~\Eqref{eq:countquery}, as a function of the log regularization strength $\lambda$. We plot the curves for (a) the Income and (b) the Employment dataset. }
    \label{fig:3}
    \vspace{-0.3in}
\end{wrapfigure}

Section \ref{sec:exp_priv_marginals} shows that PrivPGD constructs a private dataset that successfully preserves the relevant statistics of the original dataset. 
However, while differential privacy protects individual information, many applications require the protection of specific population-level statistics, i.e., one would like specific statistics from the DP-dataset $\dpdata$ to have a large \emph{mismatch} with statistics from the original dataset $D$. 
This goal is in contrast to the utility loss that tries to match certain statistics of the original dataset.
One example is census data, where it might be desirable to hide religious or sexual orientations at the subpopulation level to prevent potential discriminatory misuse of the data. Another example is the publication of rental and house sale prices, where exposing certain correlations could lead to speculative abuses in specific locales.

We now demonstrate how PrivPGD, while constructing a high-quality dataset,
 allows the protection of statistics on a population level. As an example, we consider maximizing the distance of a single linear (non-sparse) thresholding query $    \text{thrs}_{\embed}(\particles) = \frac{1}{m} \sum_i 
\mathbbm{1}_{\langle \particles_i, \theta \rangle + b >0}$ defined directly in the embedding space. We approximate this query using a differentiable sigmoid approximation and define $\hat \queryloss(Z) = \frac{1}{c + (\triangle(Z))^2}$ where $\triangle(Z)$ is the difference to a DP-estimate of the original data. We refer to the Appendix~\ref{appendix:details_domain} for details.

 We plot in Figure~\ref{fig:3} the absolute thresholding error as a function of the regularization strength $\lambda$. While the query is well approximated if no (or only little) regularization is used, we see how increasing the regularization strength increasingly protects these statistics, as the domain-specific counting query on $\dpdata$ deviates strongly from the one on the original data $D$. 
We further plot the downstream error and the absolute errors over random counting and thresholding queries (only the numerator in~\Cref{eq:countquery}). We observe that, even for larger regularization penalties, these statistics are still preserved, comparable to the unregularized case.

\section{Conclusion and future work}

We introduced PrivPGD, a novel generation method for marginal-based private data synthesis. Our approach leverages particle gradient descent, combined with techniques from optimal transport, resulting in improved performance in a large number of settings, enhanced scalability for handling numerous marginals and larger datasets, and increased flexibility for accommodating domain-specific constraints compared to existing methods.

Future work could develop regularization penalties that promote an inductive bias in PrivPGD towards more favorable solutions, similar to the maximum entropy bias in PGM. Additionally, exploring the design of domain-specific differentiable constraints and applying our method in practical scenarios presents an exciting avenue for future research.

\section{Acknowledgements}

KD was supported by the ETH AI Center and the ETH Foundations of Data Science. JA was supported by the ETH AI Center. We thank Aram Alexandre Pooladin and Guillaume Wang for insightful discussions.

%% file: sections/appendix.tex
\section{Extended Experimental Details}
\label{appendix:datasets}
\paragraph{Datasets}
We use a diverse range of real-world datasets, each with associated classification or regression tasks. Notably, our data sources include the American Community Survey (ACS) and various datasets from \cite{grinsztajn2022tree}. 

\begin{itemize}
    \item \textbf{ACS Income classification dataset (Inc.) ~\citep{ding2021retiring}}. The dataset focuses on predicting whether an individual earns more than 50,000 dollars annually. It is derived from the ACS PUMS data sample, with specific filters applied: only individuals aged above 16, those who reported working for at least 1 hour weekly in the past year, and those with a reported income exceeding 100 dollars were included. We take the data from California over a 5-year horizon and survey year 2018, with $d=11$ dimension and $n=760,157$ data points.

    \item \textbf{ACS Employment classification dataset (Emp.)~\citep{ding2021retiring}}. The dataset is designed to predict an individual's employment status. It's derived from the ACS PUMS data sample but only considers individuals aged between 16 and 90. We take the data from California over a 5-year horizon and survey year 2018, with $d=17$ dimension and $n=1,503,938$ data points.
    \item \textbf{ACS Mobility classification dataset (Mob.)~\citep{ding2021retiring}}. The goal is to determine if an individual retained the same residential address from the previous year using a filtered subset of the ACS PUMS data. This subset exclusively includes individuals aged between 18 and 35. Filtering for this age range heightens the prediction challenge, as over 90\% of the broader population typically remains at the same address from one year to the next. We take the data from California over a 5-year horizon and survey year 2018, with $d=22$ dimension and $n=318,332$ data points.
    \item \textbf{ACS Traveltime classification dataset (Tra.)~\citep{ding2021retiring}}. The objective is to predict if an individual's work commute surpasses 20 minutes using a refined subset of the ACS PUMS data. This subset is limited to those who are employed and are older than 16 years. The 20-minute benchmark was selected based on its status as the median travel time to work for the US population in the 2018 ACS PUMS data release. We take the data from California over a 5-year horizon and survey year 2018, with $d=17$ dimension and $n=667,132$ data points.
    \item \textbf{ACS Public Coverage classification dataset (Pub.)~\citep{ding2021retiring}}. The task is to predict if an individual is enrolled in public health insurance using a specific subset of the ACS PUMS data. This subset is narrowed down to individuals younger than 65 and with an income below 30,000 dollars. By focusing on this group, the prediction centers on low-income individuals who don't qualify for Medicare. We take the data from California over a 5-year horizon and survey year 2018, with $d=20$ dimension and $n=583,545$ data points.
    \item \textbf{Medical charges regression dataset (Med.)~\citep{grinsztajn2022tree}}. The dataset from the tabular benchmark, part of the ``regression on numerical features" benchmark, details inpatient discharges under the Medicare fee-for-service scheme. Known as the Inpatient Utilization and Payment Public Use File (Inpatient PUF), it provides insights into utilization, total and Medicare-specific payments, and hospital-specific charges. The dataset encompasses data from over 3,000 U.S. hospitals under the Medicare Inpatient Prospective Payment System (IPPS) framework. Organized by hospitals and the Medicare Severity Diagnosis Related Group (MS-DRG), this dataset spans from Fiscal Year 2011 to 2016. In total, it contains $n=130,452$ data point with $d=3$ features.
    \item \textbf{Black Friday regression dataset (Fri.)~\citep{grinsztajn2022tree}}. This dataset contains purchases from $n=133,456$ buyers on black Friday. Each point is described by $d=9$ features, including gender, age, occupation and marital status.
    \item \textbf{NYC Taxi Green December 2016 regression dataset (Taxi)~\citep{grinsztajn2022tree}}. The dataset, utilized in the ``regression on numerical features" benchmark from the tabular data benchmark, originates from the New York City Taxi and Limousine Commission's (TLC) trip records for the green line in December 2016. In this processed version, string datetime details have been converted to numeric columns. The goal is to predict the ``tip amount". Records exclusively from credit card payments were retained. The dataset contains $n=465,468$ points with $d=9$ features.
    \item \textbf{Diabetes 130-US dataset classification dataset (Diab.)~\citep{strack2014impact}}. The dataset encapsulates a decade (1999-2008) of $n=56,872$ clinical observations from 130 US hospitals and integrated delivery systems, comprising over $d=7$ features denoting patient and hospital results. The data was curated based on specific criteria: the record must be of an inpatient hospital admission, be associated with a diabetes diagnosis, have a stay duration ranging from 1 to 14 days, include laboratory tests, and involve medication administration.
\end{itemize}

\section{Additional Details for Section \ref{sec:domainspecific_exp}}
\label{appendix:details_domain}

We now give additional descriptions for the experiments in Section~\ref{sec:domainspecific_exp}. We approximate the thresholding function  $    \text{thrs}_{\embed}(\particles) = \frac{1}{m} \sum_i 
\mathbbm{1}_{\langle \particles_i, \theta \rangle + b>0}$  using the smooth sigmoid approximation
$
\text{s-thrsh}_{\embed}(\particles) = \frac{1}{m} \sum_i \left(1 + \exp(-\sigma (\langle \theta, \particles_i\rangle-b))\right)^{-1}
$ with $\sigma = 5.0$.  
We then split the privacy budget into two parts; the first part $\epsilon = 0.5$ and $\delta = 2 \times 10^{-6}$ is used for obtaining a DP estimate of 
$\text{s-thrsh}_{\embed}(\embed(D))$ using the Gaussian mechanim, which we denote by $\widehat{ \text{s-thrsh}}_{\embed}$. 
Moreover, we use the remaining privacy budget $\epsilon = 2.0$ and $\delta = 8 \times 10^{-6}$ to privatize all $2$-Way marginals using the Gaussian mechanism as in Algorithm~\ref{alg:mainalgo}. As a result, using the simple composition theorem (see e.g., \cite{dwork2004privacy}), the overall algorithm is then DP for $\epsilon= 2.5$ and $\delta = 10^{-5}$. 

Finally, we generate the differentially private dataset $\dpdata$ by running Algorithm~\ref{alg:pgd} with all privatized 2-Way marginals as inputs as well as the DP regularization loss
\begin{equation}
         \hat \queryloss(\particles) = \frac{0.01}{0.0001 +  ( \text{s-thrsh}_{\embed}(\particles)-\widehat{\text{s-thrsh}}_{\embed})^2}.
     \end{equation}
and regularization strength $\lambda$ reaching from $10^{-2}$ to $10$, as depicted in  Figure~\ref{fig:3}.

\section{Extended Results for Section~\ref{sec:exp_priv_marginals}}
\label{app:extended_results}

\begin{figure*}[ht!]

    \centering
  \begin{subfigure}{\linewidth}
        \centering
\includegraphics[width=0.9\textwidth]{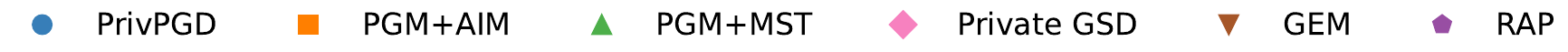}
    \end{subfigure}
    \includeBenchFigures{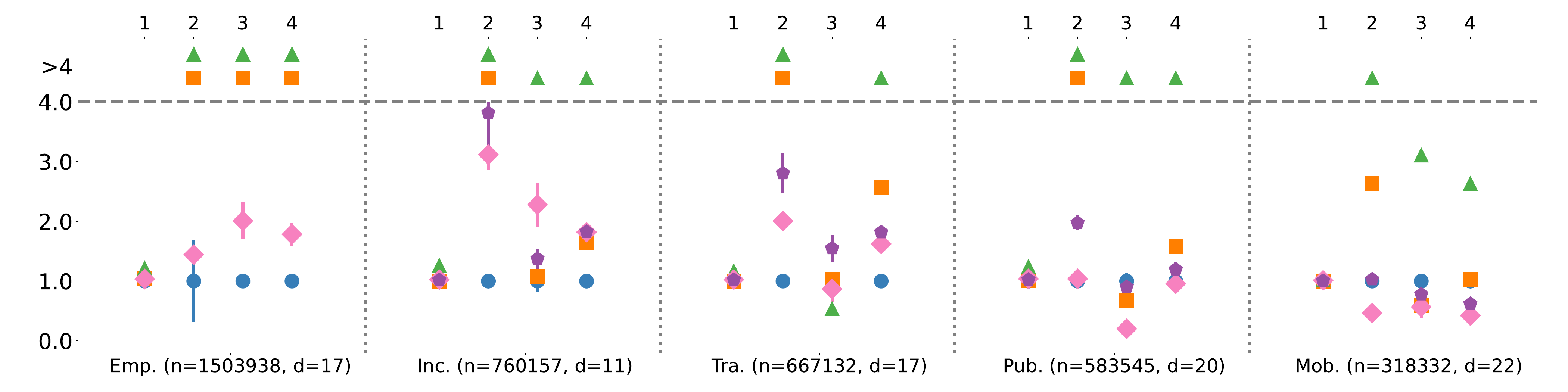}
        \centering
    \caption{\small
    Comparison of PrivPGD with all $2$-way marginals against state-of-the-art methods based on metrics from Section~\ref{subsec:expsetting}: 1) downstream error, 2) covariance error, 3) count. queries error, and 4) thresh. queries error, across 9 tabular datasets. For each method, we plot the $\log_2$ ratio of the errors, using PrivPGD's average error as the denominator, and report the mean and standard deviation over 5 runs. We cut at a log ratio of $y=3$ (dashed line) and list all methods exceeding this threshold above this line in order. We set $\epsilon = 1.0$ and $\delta = 10^{-5}$.}
    
    \label{fig:comp_1}
    \vspace{-0.2in}
\end{figure*}

\begin{figure*}[ht!]

    \centering
  \begin{subfigure}{\linewidth}
        \centering
\includegraphics[width=0.9\textwidth]{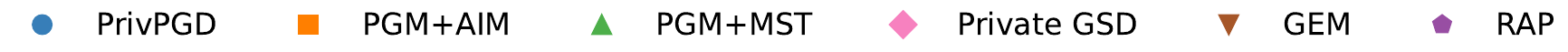}
    \end{subfigure}
    \includeBenchFigures{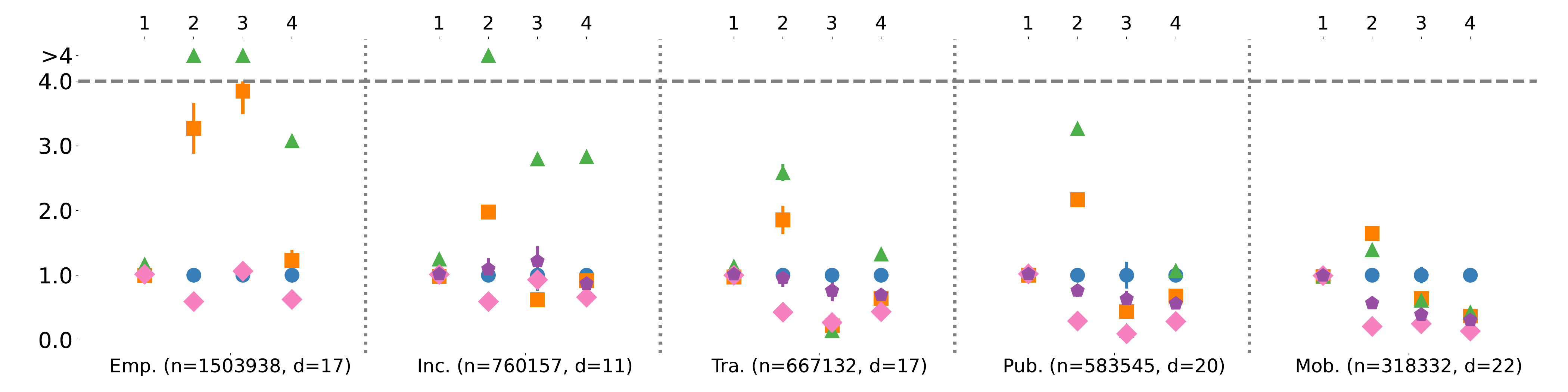}
        \centering
 \caption{\small
    Comparison of PrivPGD with all $2$-way marginals against state-of-the-art methods based on metrics from Section~\ref{subsec:expsetting}: 1) downstream error, 2) covariance error, 3) count. queries error, and 4) thresh. queries error, across 9 tabular datasets. For each method, we plot the $\log_2$ ratio of the errors, using PrivPGD's average error as the denominator, and report the mean and standard deviation over 5 runs. We cut at a log ratio of $y=3$ (dashed line) and list all methods exceeding this threshold above this line in order. We set $\epsilon = 0.2$ and $\delta = 10^{-5}$.}
    \label{fig:comp_0.2}
\end{figure*}

\begin{figure*}
    \centering
    \begin{subfigure}{\linewidth}
        \centering
        \includegraphics[width=0.9\textwidth]{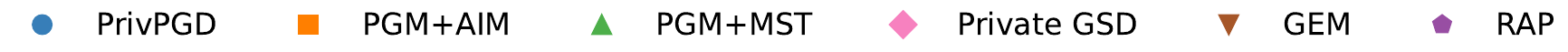}
    \end{subfigure}
    \includeDistFigures{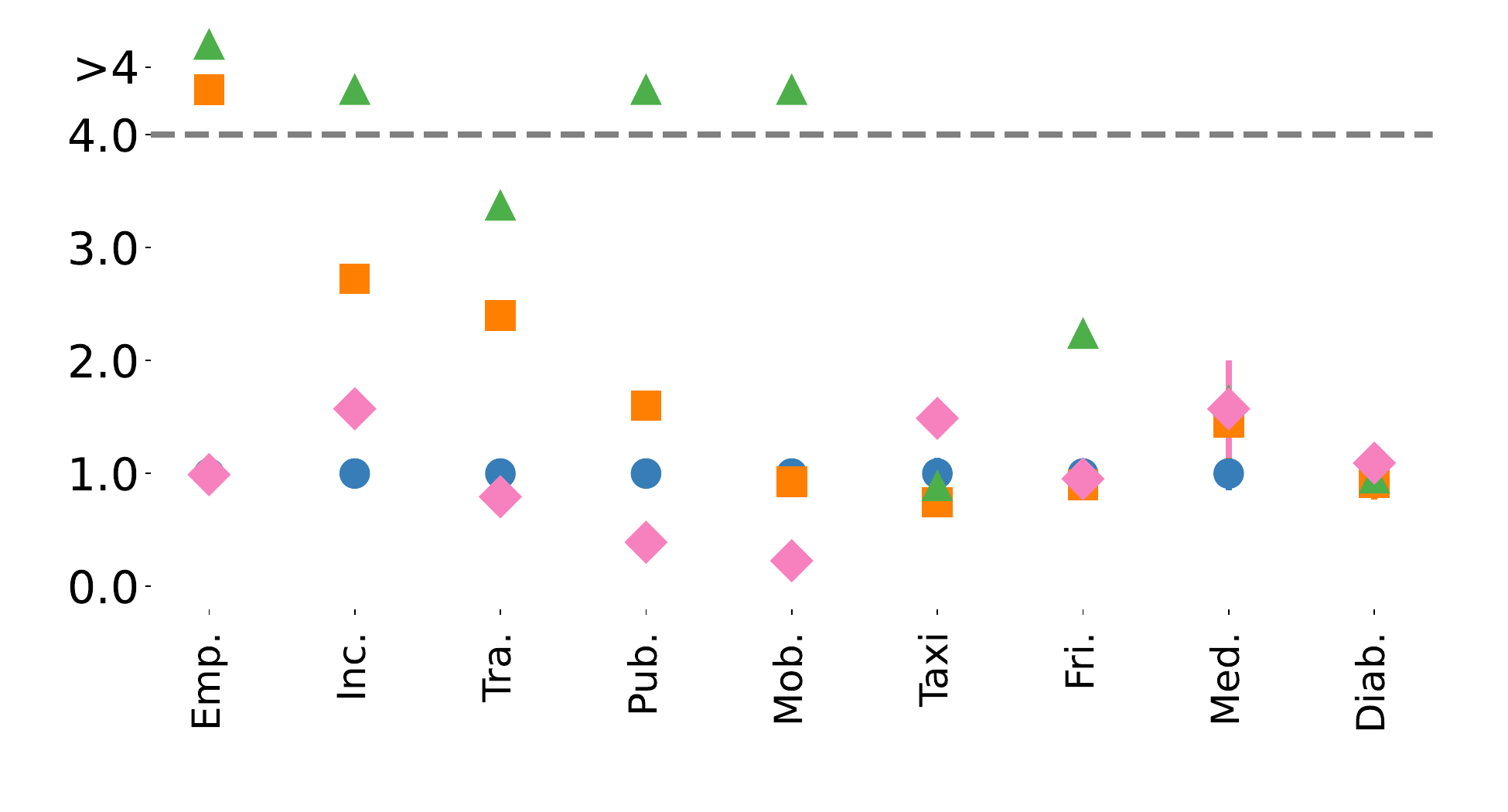}
        \centering
        \caption{\small Comparison of average $\swd_1$ distance (left) and average TV distance (right) for PrivPGD against state-of-the-art methods across 9 tabular datasets. Similar to Figure~\ref{fig:comp}, we report the mean and standard deviation (5 runs) of the $\log_2$ ratio of errors. We set $\epsilon = 1.0$ and $\delta = 10^{-5}$.}

        \label{fig:comp_sw_tv_1}
\end{figure*}

\begin{figure*}
    \centering
    \begin{subfigure}{\linewidth}
        \centering
        \includegraphics[width=0.9\textwidth]{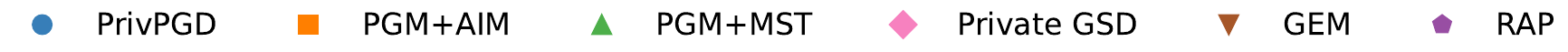}
    \end{subfigure}
    \includeDistFigures{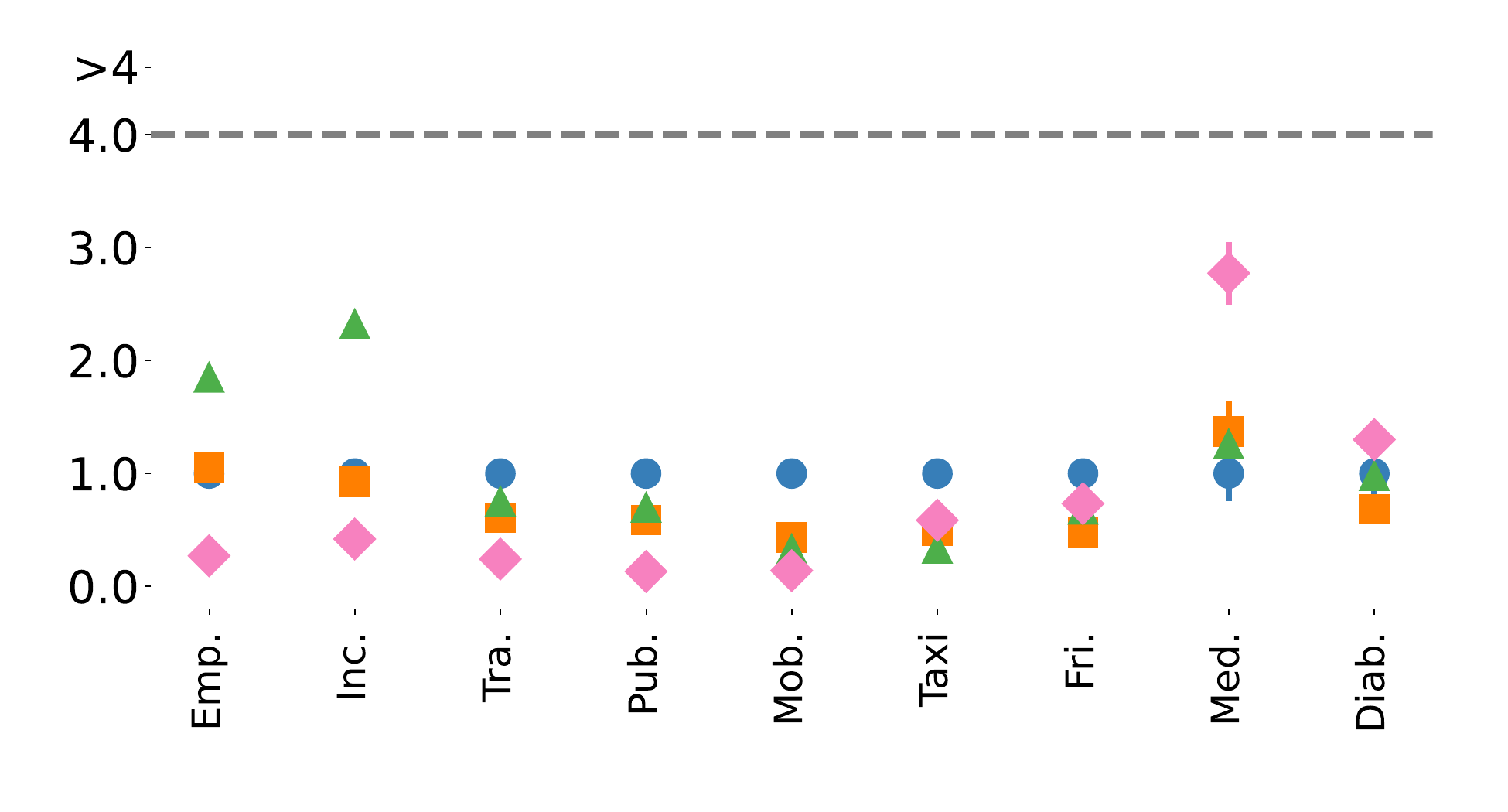}
        \centering
        \caption{\small Comparison of average $\swd_1$ distance (left) and average TV distance (right) for PrivPGD against state-of-the-art methods across 9 tabular datasets. Similar to Figure~\ref{fig:comp}, we report the mean and standard deviation (5 runs) of the $\log_2$ ratio of errors. We set $\epsilon = 0.2$ and $\delta = 10^{-5}$.}

        \label{fig:comp_sw_tv_0.2}
\end{figure*}

We replicate Figure~\ref{fig:comp} with $\epsilon=1.0$ (see Figure~\ref{fig:comp_1}) and $\epsilon=0.2$ (see Figure~\ref{fig:comp_0.2}). At smaller values of $\epsilon$, PrivPGD is more frequently outperformed by PGM, especially in combination with AIM, and also with MST at $\epsilon=0.2$, as well as more often by Private GSD. PGM-based methods benefit from a strong inductive bias towards maximum entropy solutions, which becomes particularly advantageous when noise levels are high. Nevertheless, PrivPGD still maintains highly competitive performance, especially in larger datasets and with fewer features. A similar trend is observed in the total variation and Wasserstein distance, as shown in Figures \ref{fig:comp_sw_tv_1} and \ref{fig:comp_sw_tv_0.2}. These results underscore that PrivPGD's performance edge diminishes as $\epsilon$ decreases. 

Finally, Tables \ref{table_1} to \ref{table_2} provide the detailed metrics for all our results.

\begin{table*}[ht]
\footnotesize
\centering
\begin{tabular}{cccccccc}
\toprule
dataset & inference & downstream & covariance & counting query & thresholding query & $\swd_1$ distance & TV distance \\
\midrule \multirow{4}{*}{Emp. } & PrivPGD&\textbf{0.19}/0.19 & 0.02 & \textbf{0.00087} & \textbf{0.00043} & \textbf{0.00054} & 0.021 \\
 & PGM+AIM&\textbf{0.19}/0.19 & 0.04 & 0.0025 & 0.0026 & 0.0019 & 0.026 \\
 & PGM+MST&0.33/0.19 & 0.27 & 0.012 & 0.0075 & 0.0087 & 0.1 \\
 & Private GSD&0.2/0.19 & \textbf{0.0083} & 0.0013 & 0.00073 & 0.00059 & \textbf{0.013} \\
\midrule \multirow{6}{*}{Inc. } & PrivPGD&\textbf{0.19}/0.19 & \textbf{0.001} & \textbf{0.00078} & \textbf{0.00031} & \textbf{0.00043} & \textbf{0.028} \\
 & PGM+AIM&\textbf{0.19}/0.19 & 0.014 & 0.0017 & 0.00034 & 0.0017 & 0.034 \\
 & PGM+MST&0.24/0.19 & 0.051 & 0.0077 & 0.0024 & 0.0054 & 0.11 \\
 & Private GSD&0.2/0.19 & 0.0049 & 0.0017 & 0.00083 & 0.00093 & 0.045 \\
 & GEM&0.22/0.19 & 0.047 & 0.01 & 0.0038 & 0.0099 & 0.17 \\
 & RAP&0.2/0.19 & 0.0078 & 0.0017 & 0.00056 & 0.0011 & 0.037 \\
\midrule \multirow{6}{*}{Tra. } & PrivPGD&\textbf{0.37}/0.34 & \textbf{0.0026} & \textbf{0.00065} & \textbf{0.00014} & \textbf{0.00049} & \textbf{0.027} \\
 & PGM+AIM&\textbf{0.37}/0.34 & 0.037 & 0.0023 & 0.00031 & 0.0019 & 0.035 \\
 & PGM+MST&0.46/0.34 & 0.091 & 0.0058 & 0.00032 & 0.0038 & 0.061 \\
 & Private GSD&0.38/0.34 & 0.011 & 0.0017 & 0.00037 & 0.00072 & 0.033 \\
 & GEM&0.4/0.34 & 0.048 & 0.0052 & 0.0011 & 0.0041 & 0.076 \\
 & RAP&0.38/0.34 & 0.015 & 0.0019 & 0.00047 & 0.0014 & 0.036 \\
\midrule \multirow{6}{*}{Pub. } & PrivPGD&\textbf{0.28}/0.27 & 0.06 & \textbf{0.00097} & 0.00033 & 0.00094 & 0.037 \\
 & PGM+AIM&0.32/0.27 & 0.1 & 0.0051 & 0.0012 & 0.0042 & 0.048 \\
 & PGM+MST&0.35/0.27 & 0.41 & 0.01 & 0.00074 & 0.0088 & 0.092 \\
 & Private GSD&0.3/0.27 & \textbf{0.012} & 0.0016 & \textbf{0.00015} & \textbf{0.00057} & \textbf{0.013} \\
 & GEM&0.29/0.27 & 0.035 & 0.0031 & 0.00042 & 0.0019 & 0.029 \\
 & RAP&0.29/0.27 & 0.02 & 0.0015 & 0.00054 & 0.0012 & 0.02 \\
\midrule \multirow{6}{*}{Mob. } & PrivPGD&\textbf{0.23}/0.22 & \textbf{0.0091} & 0.0014 & 0.00051 & 0.0011 & 0.056 \\
 & PGM+AIM&\textbf{0.23}/0.22 & 0.05 & 0.0029 & \textbf{0.00036} & 0.0025 & 0.035 \\
 & PGM+MST&0.24/0.22 & 0.29 & 0.0088 & 0.005 & 0.0095 & 0.096 \\
 & Private GSD&0.24/0.22 & 0.0092 & \textbf{0.0013} & 0.00068 & \textbf{0.00055} & \textbf{0.015} \\
 & GEM&0.24/0.22 & 0.063 & 0.0074 & 0.0022 & 0.006 & 0.082 \\
 & RAP&\textbf{0.23}/0.22 & 0.016 & 0.0014 & 0.00066 & 0.0013 & 0.022 \\
 \bottomrule
 \end{tabular}
 \caption{
    \small The mean of the errors from Section~\ref{subsec:expsetting}  averaged over 5 runs. For the downstream error, we additionally show the test error when training on the original private dataset. We choose $\epsilon =2.5$ and $\delta = 10^{-5}$.}
    \label{table_1}
 \end{table*}
 
 \begin{table*}[ht]
 \footnotesize
\centering
\begin{tabular}{cccccccc}
\toprule
dataset & inference & downstream & covariance & counting query & thresholding query & $\swd_1$ distance & TV distance \\
\midrule \multirow{6}{*}{Taxi } & PrivPGD&\textbf{2.3}/2.2 & \textbf{0.00048} & \textbf{0.00051} & \textbf{0.00011} & \textbf{0.00027} & 0.02 \\
 & PGM+AIM&\textbf{2.3}/2.2 & 0.0013 & 0.00066 & \textbf{0.00011} & 0.0005 & \textbf{0.016} \\
 & PGM+MST&\textbf{2.3}/2.2 & 0.0029 & 0.0011 & 0.0002 & 0.00058 & 0.028 \\
 & Private GSD&\textbf{2.3}/2.2 & 0.004 & 0.0019 & 0.00061 & 0.00093 & 0.064 \\
 & GEM&2.7/2.2 & 0.079 & 0.016 & 0.0037 & 0.018 & 0.36 \\
 & RAP&\textbf{2.3}/2.2 & 0.0065 & 0.0021 & 0.0005 & 0.0013 & 0.078 \\
\midrule \multirow{6}{*}{Fri. } & PrivPGD&\textbf{1.5}/1.4 & \textbf{0.00097} & \textbf{0.00064} & \textbf{0.00053} & \textbf{0.00033} & 0.015 \\
 & PGM+AIM&\textbf{1.5}/1.4 & 0.0014 & \textbf{0.00064} & 0.00058 & 0.00054 & \textbf{0.011} \\
 & PGM+MST&\textbf{1.5}/1.4 & 0.0057 & 0.0027 & 0.0008 & 0.0014 & 0.047 \\
 & Private GSD&1.7/1.4 & 0.0032 & 0.0015 & 0.00083 & 0.0006 & 0.021 \\
 & GEM&2.8/1.4 & 0.031 & 0.0066 & 0.0016 & 0.009 & 0.16 \\
 & RAP&\textbf{1.5}/1.4 & 0.0038 & 0.0018 & 0.00075 & 0.0011 & 0.034 \\
\midrule \multirow{6}{*}{Med. } & PrivPGD&2.2/2.1 & 0.00014 & \textbf{0.0012} & \textbf{0.00013} & \textbf{0.00036} & 0.023 \\
 & PGM+AIM&\textbf{2.1}/2.1 & 0.00066 & \textbf{0.0012} & 0.00034 & 0.00076 & \textbf{0.013} \\
 & PGM+MST&\textbf{2.1}/2.1 & 0.00061 & \textbf{0.0012} & 0.00035 & 0.0007 & 0.017 \\
 & Private GSD&2.2/2.1 & \textbf{0.00013} & \textbf{0.0012} & 0.00014 & 0.00076 & 0.02 \\
 & GEM&75/2.1 & 0.11 & 0.1 & 0.054 & 0.22 & 1.4 \\
 & RAP&4.5/2.1 & 0.0048 & 0.013 & 0.0023 & 0.007 & 0.062 \\
 \midrule
\multirow{6}{*}{Diab. }  & PrivPGD&\textbf{0.4}/0.4 & \textbf{0.0013} & \textbf{0.0013} & 0.00042 & 0.00084 & 0.028 \\
 & PGM+AIM&\textbf{0.4}/0.4 & 0.0045 & 0.0025 & 0.0015 & 0.0022 & 0.028 \\
 & PGM+MST&0.41/0.4 & 0.0047 & 0.0083 & 0.00029 & 0.0017 & 0.045 \\
 & Private GSD&0.41/0.4 & 0.0017 & 0.0017 & \textbf{0.00026} & \textbf{0.00067} & \textbf{0.026} \\
 & GEM&0.41/0.4 & 0.042 & 0.032 & 0.0084 & 0.032 & 0.23 \\
 & RAP&0.41/0.4 & 0.0034 & 0.0021 & 0.00097 & 0.0025 & 0.041 \\
\bottomrule
\end{tabular}
\caption{
    \small The mean of the errors from Section~\ref{subsec:expsetting}  averaged over 5 runs. For the downstream error, we additionally show the test error when training on the original private dataset. We choose $\epsilon =2.5$ and $\delta = 10^{-5}$.}
    \end{table*}

\begin{table*}[ht]
\footnotesize
\centering
\begin{tabular}{cccccccc}
\toprule
dataset & inference & downstream & covariance & counting query & thresholding query & $\swd_1$ distance & TV distance \\

\midrule \multirow{4}{*}{Emp. } & PrivPGD&\textbf{0.19}/0.19 & \textbf{0.0056} & \textbf{0.00077} & \textbf{0.00038} & 0.0006 & 0.027 \\
 & PGM+AIM&0.2/0.19 & 0.063 & 0.0036 & 0.004 & 0.0042 & 0.057 \\
 & PGM+MST&0.23/0.19 & 0.48 & 0.098 & 0.081 & 0.14 & 0.7 \\
 & Private GSD&0.2/0.19 & 0.008 & 0.0014 & 0.00076 & \textbf{0.00059} & \textbf{0.013} \\
\midrule \multirow{6}{*}{Inc. } & PrivPGD&\textbf{0.19}/0.19 & \textbf{0.0019} & \textbf{0.00092} & \textbf{0.00039} & \textbf{0.0006} & 0.042 \\
 & PGM+AIM&\textbf{0.19}/0.19 & 0.014 & 0.0015 & 0.00042 & 0.0016 & \textbf{0.033} \\
 & PGM+MST&0.24/0.19 & 0.052 & 0.0077 & 0.0024 & 0.0054 & 0.11 \\
 & Private GSD&0.2/0.19 & 0.0061 & 0.0017 & 0.00089 & 0.00095 & 0.046 \\
 & GEM&0.22/0.19 & 0.047 & 0.01 & 0.0037 & 0.0098 & 0.16 \\
 & RAP&\textbf{0.19}/0.19 & 0.0074 & 0.0017 & 0.00054 & 0.001 & 0.041 \\
\midrule \multirow{6}{*}{Tra. } & PrivPGD&\textbf{0.37}/0.34 & \textbf{0.0055} & \textbf{0.0011} & 0.00032 & 0.00091 & 0.05 \\
 & PGM+AIM&\textbf{0.37}/0.34 & 0.036 & 0.0027 & 0.00033 & 0.0022 & 0.042 \\
 & PGM+MST&0.44/0.34 & 0.073 & 0.0058 & \textbf{0.00017} & 0.0031 & 0.055 \\
 & Private GSD&0.38/0.34 & 0.011 & 0.0017 & 0.00028 & \textbf{0.00072} & \textbf{0.033} \\
 & GEM&0.4/0.34 & 0.048 & 0.0051 & 0.00091 & 0.004 & 0.074 \\
 & RAP&0.38/0.34 & 0.016 & 0.0019 & 0.0005 & 0.0016 & 0.041 \\
\midrule \multirow{6}{*}{Pub. } & PrivPGD&\textbf{0.28}/0.27 & \textbf{0.011} & 0.0015 & 0.00061 & 0.0014 & 0.061 \\
 & PGM+AIM&\textbf{0.28}/0.27 & 0.058 & 0.0024 & 0.00041 & 0.0023 & 0.028 \\
 & PGM+MST&0.35/0.27 & 0.29 & 0.024 & 0.0092 & 0.04 & 0.21 \\
 & Private GSD&0.29/0.27 & 0.012 & \textbf{0.0014} & \textbf{0.00012} & \textbf{0.00055} & \textbf{0.013} \\
 & GEM&0.29/0.27 & 0.031 & 0.0029 & 0.00056 & 0.0017 & 0.025 \\
 & RAP&0.29/0.27 & 0.022 & 0.0018 & 0.00055 & 0.0014 & 0.023 \\
\midrule \multirow{6}{*}{Mob. } & PrivPGD&\textbf{0.23}/0.22 & 0.021 & 0.0029 & 0.00085 & 0.0026 & 0.13 \\
 & PGM+AIM&\textbf{0.23}/0.22 & 0.054 & 0.0028 & 0.00051 & 0.0024 & 0.034 \\
 & PGM+MST&0.24/0.22 & 0.13 & 0.0077 & 0.0027 & 0.013 & 0.093 \\
 & Private GSD&0.24/0.22 & \textbf{0.0096} & \textbf{0.0012} & \textbf{0.00049} & \textbf{0.0006} & \textbf{0.016} \\
 & GEM&0.24/0.22 & 0.062 & 0.0077 & 0.0021 & 0.0062 & 0.083 \\
 & RAP&\textbf{0.23}/0.22 & 0.021 & 0.0018 & 0.00066 & 0.0017 & 0.031 \\

 \bottomrule
\end{tabular}
\caption{
    \small The mean of the errors from Section~\ref{subsec:expsetting}  averaged over 5 runs. For the downstream error, we additionally show the test error when training on the original private dataset. We choose $\epsilon =1.0$ and $\delta = 10^{-5}$.}
\end{table*}

\begin{table*}[ht]
\footnotesize
\centering
\begin{tabular}{cccccccc}
\toprule
dataset & inference & downstream & covariance & counting query & thresholding query & $\swd_1$ distance & TV distance \\
\midrule \multirow{6}{*}{Taxi } & PrivPGD&\textbf{2.3}/2.2 & \textbf{0.0013} & 0.00077 & 0.00018 & 0.00062 & 0.033 \\
 & PGM+AIM&\textbf{2.3}/2.2 & 0.0014 & \textbf{0.00057} & \textbf{0.00011} & \textbf{0.00046} & \textbf{0.018} \\
 & PGM+MST&\textbf{2.3}/2.2 & 0.003 & 0.0011 & 0.00018 & 0.00055 & 0.029 \\
 & Private GSD&2.4/2.2 & 0.0043 & 0.0018 & 0.00055 & 0.00092 & 0.066 \\
 & GEM&2.8/2.2 & 0.081 & 0.017 & 0.0045 & 0.021 & 0.36 \\
 & RAP&\textbf{2.3}/2.2 & 0.0061 & 0.0021 & 0.00052 & 0.0014 & 0.08 \\
\midrule \multirow{6}{*}{Fri. } & PrivPGD&1.6/1.4 & 0.0022 & 0.00086 & \textbf{0.00053} & 0.00068 & 0.028 \\
 & PGM+AIM&\textbf{1.5}/1.4 & \textbf{0.0016} & \textbf{0.00071} & 0.00056 & \textbf{0.00061} & \textbf{0.014} \\
 & PGM+MST&\textbf{1.5}/1.4 & 0.0058 & 0.0027 & 0.00081 & 0.0015 & 0.049 \\
 & Private GSD&1.6/1.4 & 0.0031 & 0.0015 & 0.00096 & 0.00065 & 0.024 \\
 & GEM&2.8/1.4 & 0.031 & 0.0068 & 0.0014 & 0.0092 & 0.16 \\
 & RAP&1.6/1.4 & 0.0048 & 0.002 & 0.00078 & 0.0016 & 0.046 \\
\midrule \multirow{6}{*}{Med. } & PrivPGD&2.4/2.1 & \textbf{0.00032} & 0.0015 & \textbf{0.00019} & \textbf{0.00061} & 0.03 \\
 & PGM+AIM&2.3/2.1 & 0.00076 & 0.0024 & 0.00053 & 0.00088 & 0.027 \\
 & PGM+MST&\textbf{2.1}/2.1 & 0.00088 & 0.0018 & 0.00051 & 0.001 & 0.023 \\
 & Private GSD&2.2/2.1 & 0.00041 & \textbf{0.0014} & 0.00024 & 0.00095 & \textbf{0.022} \\
 & GEM&75/2.1 & 0.11 & 0.1 & 0.054 & 0.22 & 1.4 \\
 & RAP&4.7/2.1 & 0.0051 & 0.013 & 0.0024 & 0.007 & 0.066 \\
\midrule \multirow{6}{*}{Diab. } & PrivPGD&\textbf{0.4}/0.4 & 0.0033 & 0.0026 & 0.001 & 0.002 & 0.052 \\
 & PGM+AIM&\textbf{0.4}/0.4 & \textbf{0.0031} & \textbf{0.0019} & 0.0014 & \textbf{0.0019} & \textbf{0.03} \\
 & PGM+MST&0.41/0.4 & 0.0042 & 0.0067 & \textbf{0.00056} & \textbf{0.0019} & 0.045 \\
 & Private GSD&0.41/0.4 & 0.0036 & 0.0024 & 0.00096 & 0.0022 & 0.039 \\
 & GEM&0.41/0.4 & 0.04 & 0.031 & 0.0078 & 0.03 & 0.22 \\
 & RAP&0.41/0.4 & 0.0073 & 0.0043 & 0.002 & 0.0054 & 0.071 \\
\bottomrule
\end{tabular}
\caption{
    \small The mean of the errors from Section~\ref{subsec:expsetting}  averaged over 5 runs. For the downstream error, we additionally show the test error when training on the original private dataset. We choose $\epsilon =1.0$ and $\delta = 10^{-5}$.}
    \end{table*}

\begin{table*}[ht]
\footnotesize
\centering
\begin{tabular}{cccccccc}
\toprule
dataset & inference & downstream & covariance & counting query & thresholding query & $\swd_1$ distance & TV distance \\
\midrule \multirow{4}{*}{Emp. } & PrivPGD&\textbf{0.19}/0.19 & 0.015 & 0.0022 & \textbf{0.00078} & 0.0023 & 0.11 \\
 & PGM+AIM&\textbf{0.19}/0.19 & 0.049 & 0.0027 & 0.0031 & 0.0024 & 0.035 \\
 & PGM+MST&0.23/0.19 & 0.1 & 0.0067 & 0.0047 & 0.0043 & 0.05 \\
 & Private GSD&0.2/0.19 & \textbf{0.0088} & \textbf{0.0014} & 0.00083 & \textbf{0.00063} & \textbf{0.014} \\
\midrule \multirow{6}{*}{Inc. } & PrivPGD&\textbf{0.19}/0.19 & 0.0098 & 0.0027 & 0.00086 & 0.0024 & 0.15 \\
 & PGM+AIM&\textbf{0.19}/0.19 & 0.019 & 0.0025 & \textbf{0.00053} & 0.0022 & \textbf{0.048} \\
 & PGM+MST&0.24/0.19 & 0.052 & 0.0077 & 0.0024 & 0.0056 & 0.11 \\
 & Private GSD&0.2/0.19 & \textbf{0.0058} & \textbf{0.0018} & 0.0008 & \textbf{0.001} & 0.052 \\
 & GEM&0.22/0.19 & 0.046 & 0.0099 & 0.0035 & 0.0096 & 0.16 \\
 & RAP&0.2/0.19 & 0.011 & 0.0024 & 0.001 & 0.0019 & 0.072 \\
\midrule \multirow{6}{*}{Tra. } & PrivPGD&0.38/0.34 & 0.027 & 0.0043 & 0.0012 & 0.004 & 0.19 \\
 & PGM+AIM&\textbf{0.37}/0.34 & 0.05 & 0.0028 & 0.00027 & 0.0024 & 0.048 \\
 & PGM+MST&0.44/0.34 & 0.07 & 0.0058 & \textbf{0.00018} & 0.003 & 0.057 \\
 & Private GSD&0.38/0.34 & \textbf{0.012} & \textbf{0.0019} & 0.00033 & \textbf{0.00097} & \textbf{0.041} \\
 & GEM&0.4/0.34 & 0.048 & 0.0051 & 0.00093 & 0.004 & 0.073 \\
 & RAP&0.39/0.34 & 0.026 & 0.003 & 0.00093 & 0.0033 & 0.074 \\
\midrule \multirow{6}{*}{Pub. } & PrivPGD&\textbf{0.29}/0.27 & 0.048 & 0.0058 & 0.0021 & 0.0064 & 0.24 \\
 & PGM+AIM&\textbf{0.29}/0.27 & 0.1 & 0.0039 & 0.00093 & 0.0037 & 0.043 \\
 & PGM+MST&0.31/0.27 & 0.16 & 0.0062 & 0.00031 & 0.0045 & 0.05 \\
 & Private GSD&\textbf{0.29}/0.27 & \textbf{0.014} & \textbf{0.0017} & \textbf{0.0002} & \textbf{0.00085} & \textbf{0.019} \\
 & GEM&\textbf{0.29}/0.27 & 0.044 & 0.0039 & 0.00053 & 0.0025 & 0.035 \\
 & RAP&\textbf{0.29}/0.27 & 0.037 & 0.0033 & 0.0013 & 0.003 & 0.049 \\
\midrule \multirow{6}{*}{Mob. } & PrivPGD&0.24/0.22 & 0.085 & 0.014 & 0.003 & 0.011 & 0.46 \\
 & PGM+AIM&\textbf{0.23}/0.22 & 0.14 & 0.0052 & 0.0019 & 0.0049 & 0.068 \\
 & PGM+MST&0.24/0.22 & 0.12 & 0.006 & 0.0019 & 0.0037 & 0.054 \\
 & Private GSD&0.24/0.22 & \textbf{0.018} & \textbf{0.0019} & \textbf{0.00076} & \textbf{0.0016} & \textbf{0.035} \\
 & GEM&0.24/0.22 & 0.062 & 0.0076 & 0.0018 & 0.0058 & 0.078 \\
 & RAP&0.24/0.22 & 0.048 & 0.0043 & 0.0012 & 0.0048 & 0.08 \\
 \bottomrule
 \end{tabular}
 \caption{
    \small The mean of the errors from Section~\ref{subsec:expsetting}  averaged over 5 runs. For the downstream error, we additionally show the test error when training on the original private dataset. We choose $\epsilon =0.2$ and $\delta = 10^{-5}$.}
    \end{table*}
 
\begin{table*}[ht]
\footnotesize
\centering
\begin{tabular}{cccccccc}
\toprule
dataset & inference & downstream & covariance & counting query & thresholding query & $\swd_1$ distance & TV distance \\
\midrule \multirow{6}{*}{Taxi } & PrivPGD&2.4/2.2 & 0.0055 & 0.0025 & 0.00072 & 0.0026 & 0.11 \\
 & PGM+AIM&\textbf{2.3}/2.2 & 0.0032 & \textbf{0.0011} & 0.00022 & 0.0013 & \textbf{0.037} \\
 & PGM+MST&\textbf{2.3}/2.2 & \textbf{0.003} & 0.0014 & \textbf{0.0002} & \textbf{0.00087} & 0.046 \\
 & Private GSD&2.4/2.2 & 0.0071 & 0.0022 & 0.00068 & 0.0015 & 0.087 \\
 & GEM&2.8/2.2 & 0.082 & 0.017 & 0.0042 & 0.02 & 0.36 \\
 & RAP&2.4/2.2 & 0.011 & 0.0036 & 0.00086 & 0.0036 & 0.17 \\
\midrule \multirow{6}{*}{Fri. } & PrivPGD&2.5/1.4 & 0.011 & 0.0031 & 0.00071 & 0.0031 & 0.11 \\
 & PGM+AIM&\textbf{1.5}/1.4 & \textbf{0.0048} & \textbf{0.0015} & \textbf{0.00061} & \textbf{0.0015} & \textbf{0.036} \\
 & PGM+MST&1.6/1.4 & 0.0069 & 0.003 & 0.00082 & 0.0021 & 0.065 \\
 & Private GSD&1.8/1.4 & 0.0077 & 0.0023 & 0.00088 & 0.0023 & 0.061 \\
 & GEM&2.8/1.4 & 0.031 & 0.0065 & 0.0014 & 0.0083 & 0.15 \\
 & RAP&2.1/1.4 & 0.017 & 0.0042 & 0.001 & 0.0055 & 0.12 \\
\midrule \multirow{6}{*}{Med. } & PrivPGD&3/2.1 & \textbf{0.0013} & \textbf{0.0048} & \textbf{0.00082} & \textbf{0.0024} & 0.076 \\
 & PGM+AIM&3.1/2.1 & 0.0027 & 0.013 & 0.0018 & 0.0033 & 0.12 \\
 & PGM+MST&\textbf{2.2}/2.1 & 0.0026 & 0.0053 & 0.0013 & 0.0031 & \textbf{0.055} \\
 & Private GSD&2.3/2.1 & 0.0028 & 0.0078 & 0.0017 & 0.0067 & 0.056 \\
 & GEM&76/2.1 & 0.11 & 0.1 & 0.054 & 0.21 & 1.4 \\
 & RAP&4.1/2.1 & 0.0084 & 0.018 & 0.0039 & 0.01 & 0.1 \\
\midrule \multirow{6}{*}{Diab. } & PrivPGD&0.42/0.4 & \textbf{0.01} & 0.013 & 0.0045 & 0.0082 & 0.17 \\
 & PGM+AIM&\textbf{0.41}/0.4 & 0.011 & 0.0085 & 0.0042 & \textbf{0.0056} & \textbf{0.071} \\
 & PGM+MST&\textbf{0.41}/0.4 & 0.012 & 0.01 & 0.0044 & 0.008 & 0.1 \\
 & Private GSD&\textbf{0.41}/0.4 & 0.014 & \textbf{0.0083} & \textbf{0.0041} & 0.011 & 0.12 \\
 & GEM&\textbf{0.41}/0.4 & 0.043 & 0.033 & 0.0076 & 0.032 & 0.23 \\
 & RAP&0.42/0.4 & 0.021 & 0.014 & 0.0061 & 0.018 & 0.19 \\
\bottomrule
\end{tabular}
 \caption{
    \small The mean of the errors from Section~\ref{subsec:expsetting}  averaged over 5 runs. For the downstream error, we additionally show the test error when training on the original private dataset. We choose $\epsilon =0.2$ and $\delta = 10^{-5}$.}
    \label{table_2}
    \end{table*}